\documentclass{article} 
\usepackage{collas2022_conference,times}

\usepackage{amsmath,amsfonts,bm}

\def\eqref#1{equation~\ref{#1}}

\def\1{\bm{1}}

\DeclareMathAlphabet{\mathsfit}{\encodingdefault}{\sfdefault}{m}{sl}
\SetMathAlphabet{\mathsfit}{bold}{\encodingdefault}{\sfdefault}{bx}{n}

\usepackage[unicode]{hyperref}
\hypersetup{
    colorlinks=true,
    linkcolor=red,
    filecolor=magenta,      
    urlcolor=blue,
    citecolor=purple,
    pdftitle={Overleaf Example},
    pdfpagemode=FullScreen,
    }

\usepackage[utf8]{inputenc} 
\usepackage[T1]{fontenc}    
\usepackage{xurl} 
\usepackage{booktabs}       
\usepackage{amsfonts}       
\usepackage{nicefrac}       
\usepackage{microtype}      
\usepackage{xcolor}         
\usepackage{amsmath,amsthm,amssymb}
\usepackage{mathtools}
\usepackage{dsfont}
\usepackage{graphicx}
\usepackage{caption}
\usepackage{subcaption}
\usepackage{adjustbox}

\usepackage{xparse}
\DeclareDocumentCommand \expectation { o m } {%
  \ensuremath{\mathbb{E}%
  \IfValueTF {#1} {%
    _{#1} \left[ #2 \right]%
  }{%
    \left[ #2 \right]%
  }%
  }\xspace%
}

\usepackage{enumitem}
\usepackage{wrapfig}
\usepackage{float}
\usepackage{microtype}

\newcommand{\MDP}{\mathcal{M}}
\newcommand{\Actions}{\mathcal{A}}
\newcommand{\States}{\mathcal{S}}
\newcommand{\Reward}{R}
\newcommand{\Transition}{T}

\newcommand{\IIWA}{\texttt{IIWA}}
\newcommand{\Jaco}{\texttt{Jaco}}
\newcommand{\Panda}{\texttt{Panda}}
\newcommand{\Kinova}{\texttt{Gen3}}

\newcommand{\boxObject}{\texttt{box}}
\newcommand{\hollowBox}{\texttt{hollow\_box}}
\newcommand{\dumbbell}{\texttt{dumbbell}}
\newcommand{\plate}{\texttt{plate}}

\newcommand{\objectWall}{\texttt{object\_wall}}
\newcommand{\objectDoor}{\texttt{object\_door}}
\newcommand{\goalWall}{\texttt{goal\_wall}}
\newcommand{\noObstacle}{\texttt{no\_obstacle}}

\newcommand{\pickPlace}{\texttt{pick-and-place}}
\newcommand{\breakablePickPlace}{\texttt{pick-{\allowbreak}and-{\allowbreak}place}}
\newcommand{\push}{\texttt{push}}
\newcommand{\shelf}{\texttt{shelf}}
\newcommand{\trashcan}{\texttt{trash\_can}}

\newcommand{\robosuite}{\texttt{robosuite}}

\newcommand*{\medcup}{\mathbin{\raisebox{1.5pt}{\scalebox{0.8}{\ensuremath{\bigcup}}}}}%

\newcommand{\benchmark}{CompoSuite}
\renewcommand{\paragraph}[1]{{\bf #1}~~}

\title{\benchmark{}: A Compositional\\ Reinforcement Learning Benchmark}

\newcommand\blfootnote[1]{%
  \begingroup
  \renewcommand\thefootnote{}\footnote{#1}%
  \addtocounter{footnote}{-1}%
  \endgroup
}

\author{Jorge A. Mendez{\normalfont $^*$,} Marcel Hussing{\normalfont $^*$,} Meghna Gummadi{\normalfont, and} Eric Eaton\\
    Department of Computer and Information Science, 
    University of Pennsylvania\\
    \texttt{\{mendezme,mhussing,meghnag,eeaton\}@seas.upenn.edu}}

\collasfinalcopy 

\begin{document}

\maketitle
\begin{abstract} \label{sec:abstract}

\blfootnote{\hspace{-0.4em}$^*$The two first authors contributed equally to this work.}We present \benchmark{}, an open-source simulated robotic manipulation benchmark for compositional multi-task
reinforcement learning (RL). Each \benchmark{} task requires a particular \textit{robot} arm to manipulate one individual \textit{object} to achieve a task \textit{objective} while avoiding an \textit{obstacle}. This compositional definition of the tasks endows \benchmark{} with two remarkable properties. First, varying 
the robot/object/objective/obstacle elements leads to hundreds of RL tasks, each of which requires a meaningfully different behavior. Second, RL approaches can be evaluated specifically for their ability to learn the compositional structure of the tasks. 
This latter capability to functionally decompose problems would enable intelligent agents to identify and exploit commonalities between learning tasks to handle large varieties of highly diverse problems. 
We benchmark existing single-task, multi-task, and compositional learning algorithms on various training settings, and assess their capability to compositionally generalize to unseen tasks. Our evaluation exposes the shortcomings of existing RL approaches with respect to compositionality and opens new avenues for  investigation.

\end{abstract}

\section{Introduction} \label{sec:intro}

Compositionality is ubiquitous in artificial and biological computational systems: it arises in natural language, logical reasoning, and software programs, among others. Artificial intelligence (AI) has leveraged composititionality from early work on hierarchical planning~\citep{sacerdoti1974planning} and logic-based reasoning~\citep{doyle1979truth}, to modern learning-based techniques like neural module networks~\citep{andreas2016neural} and skill discovery~\citep{konidaris2009skilldiscovery}. The ability to decompose a complex problem into easier subproblems, such that the solutions to these subproblems can be combined into an overall solution, could increase the capabilities of learning agents. First, the learning problem itself might become easier, due to the ease of learning each subproblem. In particular, if the agent is learning multiple tasks that share common subproblems, it could amortize the cost of discovering the decomposition across the multiple tasks. Moreover, the agent could quickly solve new tasks by discovering which components are suitable to these new tasks---in the most extreme case, if the agent is informed about which components are needed (e.g., in the form of a task descriptor), then it could achieve \textit{zero-shot compositional generalization} without requiring any data from the new tasks.

Despite the appeal of these ideas, few reinforcement learning (RL) efforts have sought to leverage compositional properties of the environment to generalize to \textit{unknown} combinations of \textit{known} components. In this work, we present \benchmark{}\footnote{\url{https://github.com/Lifelong-ML/CompoSuite}}, a benchmark for compositional RL that exploits the compositionality of robot learning tasks to evaluate the compositional capabilities of learning agents. We follow the functional composition formulation, which decomposes the learning problem into subproblems whose outputs become the inputs to other subproblems~\citep{mendez2022modular}, akin to the decomposition of programs for solving robot tasks into software modules for sensing, planning, and acting.

\benchmark{} comprises \textit{hundreds} of RL tasks,  each made up of four components: robot arm, obstacle, object, and task objective. For example, one task requires an IIWA arm to circumvent a wall, pick up a dumbbell, and place it in a bin. Another task instructs a Jaco arm to traverse a doorway, pick up a plate, and insert it into a shelf. If a learner appropriately decomposes its solutions to these two problems into functional components, it could reuse the IIWA motor module in place of the Jaco motor module to solve the plate-on-shelf task without any experience with the IIWA arm on that task. More generally, multi-task and continual RL approaches can be evaluated on \benchmark{} for their ability to handle large numbers of highly varied tasks. This is in stark contrast to most existing multi-task RL benchmarks, which are typically limited to at most a few dozen RL tasks: we offer an order of magnitude more tasks, enabling the study of multi-task and continual RL at scale. The main contributions of this work include: 
\vspace{-0.5em}
\begin{itemize}[leftmargin=*,noitemsep,topsep=0pt,parsep=0pt,partopsep=0pt]
    \item \benchmark{}, a benchmark of $256$ compositional simulated robotic manipulation tasks with distinct optimal behaviors.
    \item Evaluation schemes and metrics for reproducible evaluation of future approaches under \benchmark{}.
    \item An evaluation of single-task, end-to-end multi-task, and modular multi-task agents on various settings under \benchmark{}, showing that these existing approaches fall short of fully exploiting the compositional properties of \benchmark{}.
\end{itemize}

\section{Related work} \label{sec:related}

\paragraph{Composition in supervised learning}
The majority of work on learning compositional representations has been in the supervised setting. One recently popular idea has been to learn a separate neural net module for each task component, encoding the composition directly in the architecture. Approaches in this setting assume that the agent is given access to the ground-truth compositional graph~\citep{andreas2016neural, hudson2018compositional}, that it must learn the graph directly from data~\citep{rosenbaum2018routing,alet2018modular,chang2018automatically}, or that it must learn a soft approximation to the graph~\citep{kirsch2018modular,meyerson2018beyond}. These works have shown not only that modularity can increase data efficiency, but also that modular agents are better able to generalize to new combinations of known task components. In the continual or lifelong learning setting, most modular approaches assume that modules can be trained on a single task and used to generalize to future tasks~\citep{reed2016neural,fernando2017pathnet,valkov2018houdini,veniat2021efficient}. More recent techniques instead follow a continual learning process that continues to improve modules with knowledge from future tasks~\citep{mendez2021lifelong,ostapenko2021continual}.

\paragraph{Composition in reinforcement learning} 
Compositionality has also been studied in hierarchical RL in the form of \textit{temporal} abstractions~\citep{sutton1999between}. This has led to approaches that automatically learn skills that can be chained in sequence to execute complex behaviors~\citep{konidaris2009skilldiscovery,bacon2017option,Sharma2020Dynamics-Aware}, and that can be transferred to new tasks by learning a new mechanism to combine known skills~\citep{florensa2017stochastic}. Another form of hierarchical RL instead leverages state abstractions to improve learning efficiency by learning the policy on a compressed representation of the state~\citep{dayan1993feudal, dietterich2000hierarchical,vezhnevets2017feudal, abel2018state}. 
Other recent works have composed policies to solve new tasks that are logical compositions of known tasks~\citep{todorov2009compositionality, barreto2018transfer, haarnoja2018composable, van2019composing, nangue2020boolean}.

Functional composition has received far less attention in RL. \citet{devin2017learning} combined neural net modules to solve robotics tasks, and others have automatically discovered the decomposition of a policy into modules~\citep{goyal2021recurrent,mittal2020learning,yang2020multi}. Most recently, \citet{mendez2022modular} formalized functional compositionality in RL, and demonstrated that it improves sample efficiency and compositional generalization in multi-task and continual RL.

\paragraph{Existing benchmarks} 
The development of large-scale, standardized benchmarks has been key to the acceleration of deep learning research (e.g., ImageNet; \citealp{deng2009imagenet}). Efforts to create equivalent advancements in deep RL have led to popularly used evaluation domains in both discrete-~\citep{bellemare2013arcade,vinyals2017starcraft} and continuous-action~\citep{brockman2016openai,tunyasuvunakool2020dmcontrol} settings. However, these benchmarks are restricted to single-task training---each task is designed to be learned \textit{in isolation}. Consequently, work in multi-task and continual RL has resorted to ad hoc evaluation settings, slowing down progress. Recent efforts have sought to bridge this gap by creating evaluation domains with multiple tasks that share a common structure that is (hopefully) transferable across the tasks. One example varied dynamical system parameters (e.g., gravity) of continuous control tasks~\citep{henderson2017multitask}. Other work created a grid-world evaluation domain with tasks of progressive difficulty~\citep{chevalier-boisvert2018babyai}. In the continual learning setting, a recent benchmark was proposed based on multi-agent coordination~\citep{nekoei2021continuous}. In the context of robotics, large sets of tasks have recently been created for evaluating multi-task, continual, and meta-learning algorithms~\citep{yu2019meta,james2020rlbench, wolczyk2021continual}.

Despite this recent progress, it remains unclear exactly \textit{what} an agent can transfer between tasks in these benchmarks, and so existing algorithms are typically limited to transferring neural net parameters in the hopes that they discover reusable information. Unlike these existing benchmarks, \benchmark{} is designed around a set of shared components, such that the commonalities across tasks are precisely understood, following equivalent efforts from the supervised setting~\citep{bahdanau2018systematic, lake2018generalization, sinha2020evaluating,vedantam2021curi}. A similar benchmark was recently proposed for evaluating temporal (instead of functional) compositionality~\citep{gur2021environment}. Another related benchmark procedurally created robotics tasks by varying dynamical parameters to study causality in RL~\citep{ahmed2021causalworld}, but considered a single robot arm and continuous variations in the physical properties of objects. 

\begin{figure}
\captionsetup[subfigure]{aboveskip=1pt}
\captionsetup{aboveskip=2pt}
    \begin{subfigure}[b]{0.25\textwidth}
        \centering
            \captionsetup{width=0.95\linewidth,justification=centering,singlelinecheck=false}
            \includegraphics[width=0.95\linewidth, trim={0cm 5cm 0cm 4cm}, clip]{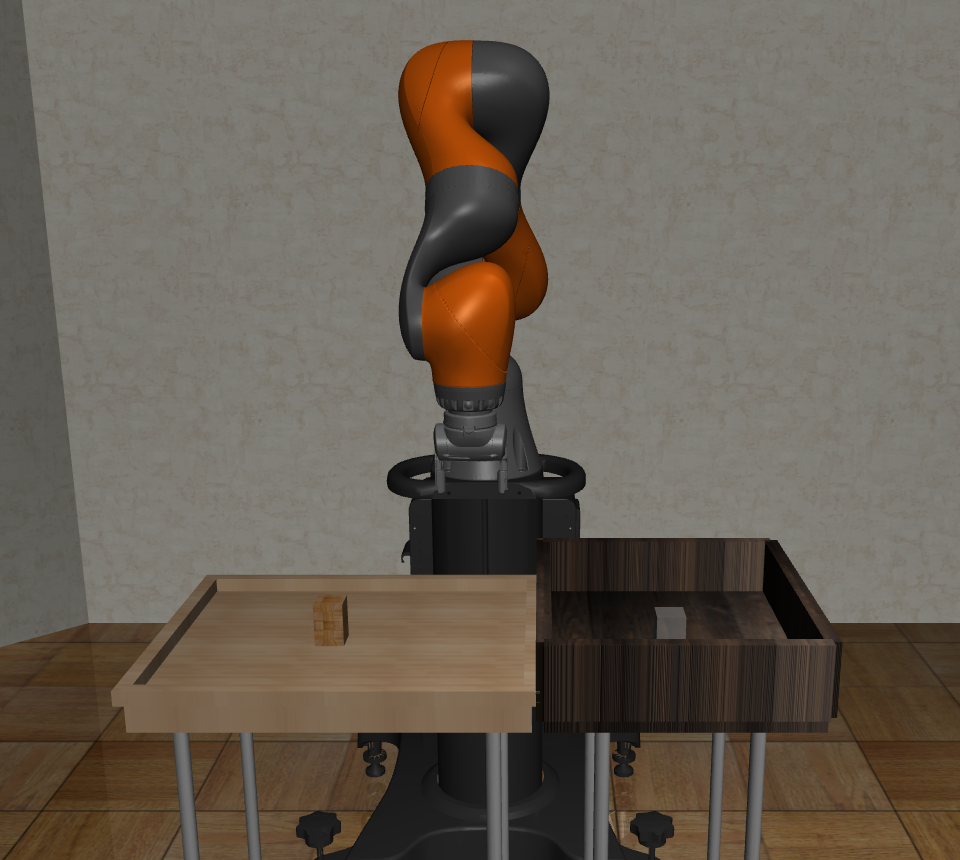}
            \caption*{{\small$\langle$\IIWA{}, \boxObject{}, \noObstacle{}, \pickPlace{}$\rangle$}}
            \label{fig:arena1}
    \end{subfigure}%
    \begin{subfigure}[b]{0.25\textwidth}
        \centering
            \captionsetup{width=0.95\linewidth,justification=centering,singlelinecheck=false}
            \includegraphics[width=0.95\linewidth, trim={0cm 5cm 0cm 4cm}, clip]{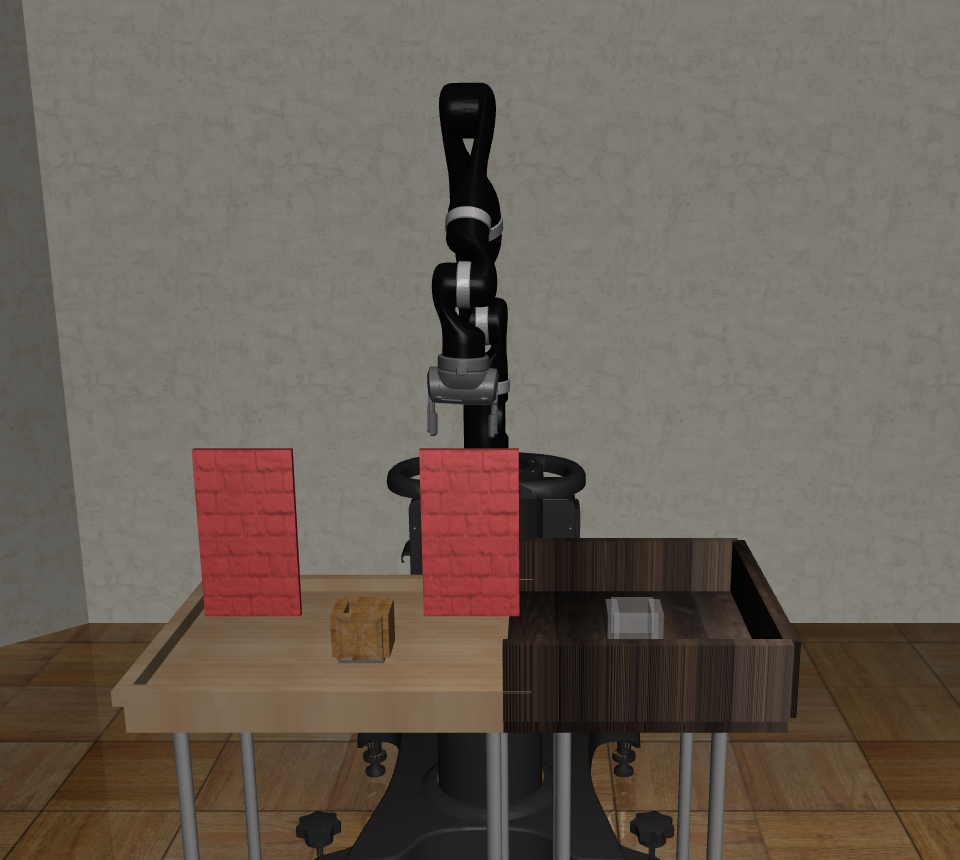}
            \caption*{{\small$\langle$\Jaco{}, \hollowBox{}, \objectDoor{}, \push{}$\rangle$}}
    \end{subfigure}%
    \begin{subfigure}[b]{0.25\textwidth}
        \centering
            \captionsetup{width=0.95\linewidth,justification=centering,singlelinecheck=false}
            \includegraphics[width=0.95\linewidth, trim={0cm 5cm 0cm 4cm}, clip]{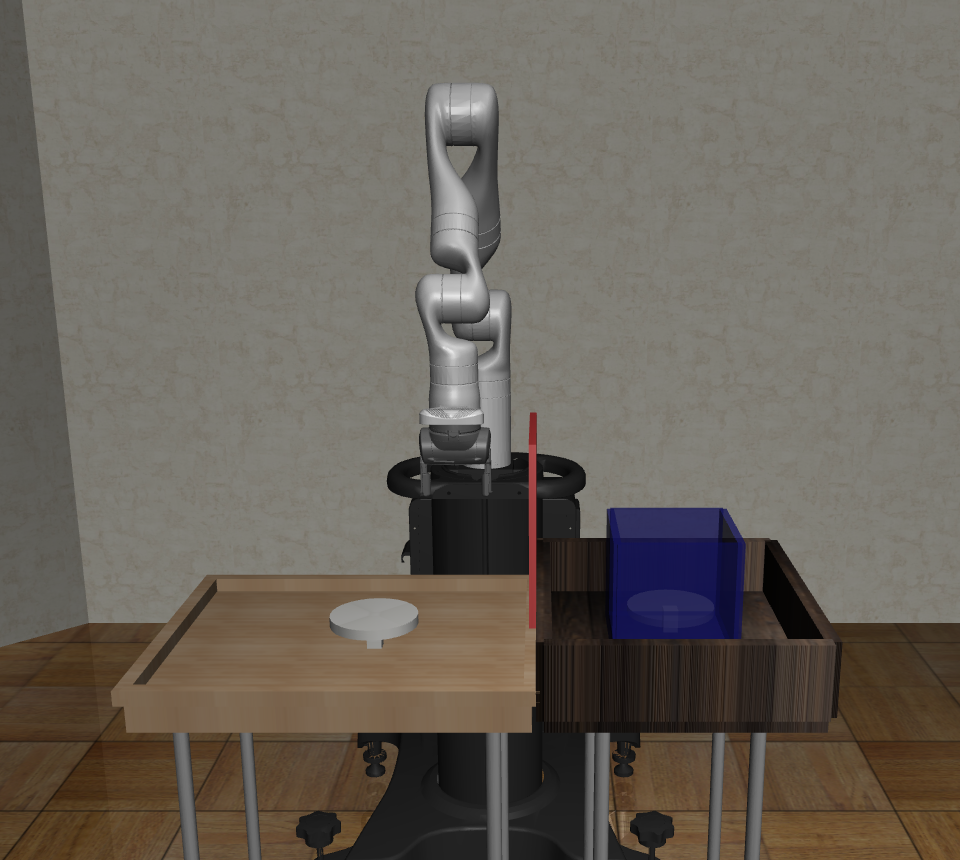}
            \caption*{{\small$\langle$\Kinova{}, \plate{}, \goalWall{}, \trashcan{}$\rangle$}}
    \end{subfigure}%
    \begin{subfigure}[b]{0.25\textwidth}
        \centering
            \captionsetup{width=0.95\linewidth,justification=centering,singlelinecheck=false}
            \includegraphics[width=0.95\linewidth, trim={0cm 5cm 0cm 4cm}, clip]{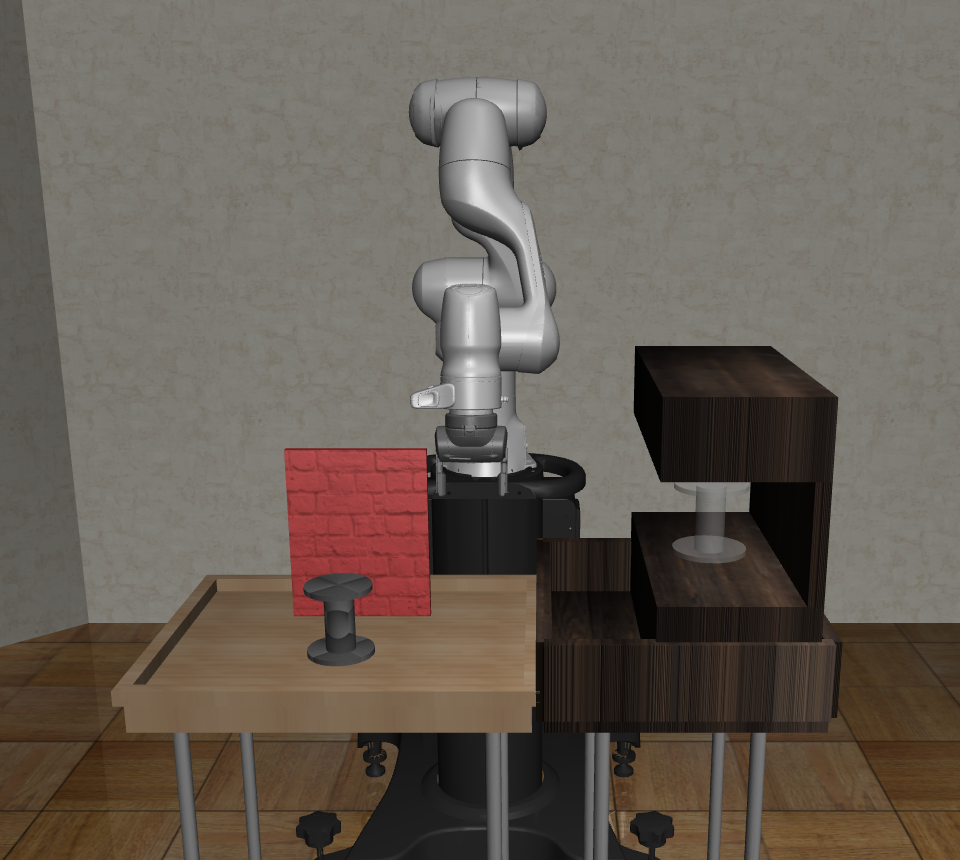}
            \caption*{{\small$\langle$\Panda{}, \dumbbell{}, \objectWall{}, \shelf{}$\rangle$}}
    \end{subfigure}
    \caption{Initial conditions of four \benchmark{} tasks, containing all elements of each compositional axis. Robots: \IIWA{}, \Jaco{}, \Kinova{}, and \Panda{}. Objects: \boxObject{}, \hollowBox{}, \plate{}, and \dumbbell{}. Obstacles: \noObstacle{}, \objectDoor{}, \goalWall{}, \objectWall{}. Objectives: \pickPlace{}, \push{}, \trashcan{}, \shelf{}.}\label{fig:arenas}
\end{figure}

\section{Background on functionally compositional RL} \label{sec:problem}

Many complex problems can be solved by considering smaller components of the problem separately and combining their solutions. This type of knowledge composition has long been considered a promising direction in AI. A recently proposed approach to composition is the notion of functional compositionality of RL tasks~\citep{mendez2022modular}. 

An RL problem is given by a Markov decision process (MDP) $\MDP=\langle \States, \Actions, \Reward, \Transition, \gamma \rangle$, where $\States$ is the set of states, $\Actions$ is the set of actions, $\Reward(s, a)$ is the reward function, $\Transition(s' \mid s, a)$ is the transition function, and $\gamma$ is the reward discount factor. Solving the MDP involves finding the policy $\pi^*$ that maximizes the expected returns $E_{\pi}[\sum_{t=0}^{\infty} \gamma^{t} \Reward(s_{t}, a_{t})]$.

In multi-task RL, the agent is faced with a set of MDPs $\mathcal{T} = \{\MDP_1, \MDP_2,\ldots,\MDP_N \}$, and its goal is to solve all the MDPs. In particular, functionally compositional MDPs are assumed to be compositions of different subsets from a set of $k$ shared subproblems $\mathcal{F}=\{F_1, F_2,\ldots,F_k\}$. To solve the MDPs, the agent could find the set of solutions to these subproblems $M = \{m_1, ..., m_k\}$, such that any MDP $\MDP_i\in\mathcal{T}$ can be solved by combining the correct subproblem solutions from $M$. In the context of robotics, each of these subproblems can be thought of as a function in a processing pipeline, such as obstacle detection, object recognition, planning, and control. This idea can be formalized as a graph $G = \{\mathcal{V}, \mathcal{E}\}$, where the vertices $\mathcal{V} = \mathcal{F} \medcup \breve{\States} \medcup \breve{\Actions}$ correspond to subproblem solutions $\mathcal{F}$, state spaces $\breve{\States} = \mathrm{unique}\left(\left\{\States^{(1)}, \ldots, \States^{(N)}\right\}\right)$, and action spaces $\breve{\Actions} = \mathrm{unique}\left(\left\{\Actions^{(1)}, \ldots, \Actions^{(N)}\right\}\right)$. Then, any MDP $\MDP_i\in\mathcal{T}$ is specified as a pair of nodes $(\States_i, \Actions_i)$ in the graph, and the optimal policy $\pi^*_i$ can be represented as a path on the graph.

A natural solution to this problem is to learn individual functions (e.g., neural net modules) to solve each subproblem $m_i\in M$. However, the problem setting is not restricted to this specific choice.
Moreover, while this notion is related to hierarchical RL, it departs from \textit{temporal} compositions and studies general \textit{function} compositions.

\section{The \benchmark{} benchmark for compositional RL} \label{sec:benchmark}

\benchmark{} is a simulated robotic manipulation benchmark designed to study the ability of RL algorithms to learn functional decompositions of the solutions to the tasks, yet more broadly applicable to multi-task and continual RL. The key idea is to build the tasks compositionally, so that 1)~we can create combinatorially many (distinct) tasks, and 2)~tasks are explicitly compositionally related.  
Sampled tasks are illustrated in Figure~\ref{fig:arenas} and all tasks are shown in Appendix~\ref{app:VisualizationOfAllTasks}. 

\subsection{Task design}

\benchmark{} is implemented on top of \robosuite{}~\citep{yuke2020robosuite}, a framework for the design of new simulated robotics environments in MuJoCo~\citep{todorov2012mujoco}. Concretely, \benchmark{} is built around four compositional axes, which represent common modules that typically make up robotic manipulation programming pipelines: object grasp-pose detection, obstacle avoidance, task planning, and low-level motor control. There are four elements of each type (i.e., for each axis), so that combining them yields a total of $256$ tasks; this represents the largest discrete set of tasks in a multi-task RL benchmark to date, yet remains computationally feasible. Within each axis, elements are designed such that a policy that succeeds at one task is very unlikely to succeed at another task---and the optimal policy for one task is even less likely to be optimal for another. To focus on the property of compositionality, variations within each axis are discrete, such that an agent can not trivially interpolate between the elements within one axis.
Each environment contains two bins: one for objects and one for targets. This standardization encourages the agents to find the commonalities between the tasks. The reward functions are crafted to facilitate learning each individual task. 

\subsubsection{Task components}
\paragraph{Robots} 
As the first axis of \benchmark{}, we use simulated versions of commercially available robotic manipulators: KUKA's \IIWA{}, Kinova's \Jaco{}, Franka's \Panda{}, and Kinova's \Kinova{}. These manipulators vary in sizes, kinematic configurations, and position and torque limits, leading to semantic discrepancies between their observations and actions that require the agent to specialize its control policy for each arm. Consequently, a policy that works on one robot arm cannot be directly applied to another arm. To ensure compatibility with existing multi-task RL methods
, we use arms with seven degrees of freedom (7-DoF).  All arms use the Rethink Robotics two-finger gripper to manipulate objects.

\paragraph{Objects} We next consider four objects of distinct shapes that require orthogonal grasping orientations. 
The \boxObject{} is a cuboid that can be picked up from the top. The \hollowBox{} resembles an open package, with a size sufficiently large that the gripper cannot grasp it by both sides like the \boxObject{}, and must instead grip one of its edges. The \dumbbell{} is placed upright, and its weights are larger than the gripper, and so it can only be grasped horizontally by the bar. The \texttt{plate}'s diameter is also greater than the gripper size and therefore can only be grasped horizontally by the edge.

\paragraph{Obstacles} 
The third axis of variation in \benchmark{} is a set of four obstacles that block off distinct areas for trajectory planning.  
The \objectWall{} is a brick wall placed between the robot and the object, while the \objectDoor{} is a similarly placed doorway between two brick walls. These two obstacles require avoiding opposite regions of the space while reaching for the object. The \goalWall{} is also a brick wall, but is placed between the left and right bins, blocking the direct path to goal after grasping the object. Additionally, we consider tasks with \noObstacle{}.

\paragraph{Task objectives} The final compositional axis is a set of different task objectives, each of which requires a unique sequence of steps for successful completion. The objectives are to \pickPlace{} an object into the right bin, \push{} the object from the left to the right bin, drop the object into a \trashcan{}, and place the object on a \shelf{}. 

Thanks to combinatorial explosion, there are $256$ possible combinations of these components, leading to a set of $256$ highly varied tasks. By design, each task requires a unique policy, but we know exactly how tasks relate to one another, enabling researchers to extract insights about the kind of compositionality that deep RL methods exhibit.

\subsubsection{Observation and action spaces}
\label{sec:observationSpace}

The observation space is split into the following factors, tied to the task components described in the previous section:
\vspace{-0.5em}
\begin{itemize}[leftmargin=*,noitemsep,topsep=0pt,parsep=0pt,partopsep=0pt]
\item \textit{Robot observation}~~The proprioceptive portion of the observation space includes the sine and cosine of the robot's joint positions, its joint velocities, end effector pose, finger positions, and finger velocities.

\item \textit{Object observation}~~The agent observes both the absolute position and orientation of the object in world coordinates, as well as its position and orientation with respect to the robot's end effector. Note that this observation deliberately does not give away any information that distinguishes objects from one another (e.g., their geometric properties).

\item \textit{Obstacle observation}~~The agent also observes the absolute and relative positions and orientations of the obstacles. Similarly, this does not give away what the free space of the environment is (e.g., \objectWall{} and \objectDoor{} are always placed in the same location, but they block off opposite parts of the space).

\item \textit{Goal observation}~~The agent is also given the absolute and relative position and orientation of the goal, as well as the relative position of the goal with respect to the object. However, for simplicity, \pickPlace{}, \trashcan{}, and \shelf{} tasks are considered solved at any arbitrary location in the target region (e.g., the right bin or the shelf). 

\item \textit{Task observation}~~The agent may also be given access to a multi-hot indicator that identifies each of the components of the task (i.e., the robot, object, obstacle, and objective). This is used as a task descriptor for multi-task training.
\end{itemize}

The action space is eight-dimensional, with the first seven dimensions providing target joint angles. 
Under the hood, a proportional-derivative (PD) controller executes the motor commands that follow the joint positions provided by the agent.
The eighth dimension is a binary action that indicates whether the gripper should be open or closed. 

\subsubsection{Reward functions}
\label{sec:RewardFunctions}

While \benchmark{} supports sparse rewards for successful completion, this leads to an extremely hard exploration problem. Consequently, to isolate the problem of multi-task compositional learning, we provide a crafted reward that encourages exploration in stages, such that each stage leads the agent to a state that is closer to task completion.

During the initial \textbf{reach} stage, the agent is rewarded for reducing the distance from the gripper to the object. This stage terminates once the agent \textbf{grasps} the object, which gives a binary reward. These two initial stages are common to all objectives. In all tasks except for \push{}, the agent is next rewarded for \textbf{lifting} the object up to a given height. In the case of \shelf{} tasks, the agent is then encouraged to \textbf{align} the gripper with the horizontal plane, facing the shelf. The next stage rewards the agent for \textbf{approaching} the right bin (or the goal, in \push{} tasks) based on the horizontal distance. In \pickPlace{} tasks, the reward then encourages the agent to \textbf{lower} the object down to the bin. In \trashcan{} tasks, the agent is instead rewarded for \textbf{dropping} the object while above the trash can with a binary reward. 

The final stage is a binary \textbf{success} reward. \pickPlace{} tasks succeed if the object is in the bin and the robot is near the object; this latter constraint differentiates \pickPlace{} and \trashcan{} tasks. \push{} tasks are solved if the object is near the goal location. The agent succeeds on \trashcan{} tasks if the object is inside the trashcan and the gripper is \textit{not}. The success criterion for \shelf{} tasks is that the object is on the shelf. 

The maximum possible reward is $\Reward=1$ and is only attained upon successfully executing the task. Table~\ref{tab:stagedRewards} summarizes the stages of each task objective, and precise formulas for the task objective rewards are included in Appendix~\ref{app:RewardFunctions}.

\begin{table}[htb]
    \centering
    \caption{Reward stages per task objective. The agent is encouraged to solve each stage before moving to the next.}
    \label{tab:stagedRewards}
    \vspace{-0.5em}
    \begin{tabular}{l|l@{}l@{}l@{}l@{}l@{}l@{}l}
        Task & \multicolumn{7}{c}{Stages}\\
        \hline
        \pickPlace{} & reach $\rightarrow\,\,$& grasp $\rightarrow\,\,$& lift $\rightarrow\,\,$& & approach $\rightarrow\,\,$& lower $\rightarrow\,\,$ & success  \\
        \push{} & reach $\rightarrow\,\,$& grasp $\rightarrow\,\,$& & & approach $\rightarrow\,\,$ & & success \\
        \trashcan{} & reach $\rightarrow\,\,$& grasp $\rightarrow\,\,$& lift $\rightarrow\,\,$& & approach $\rightarrow\,\,$& drop $\rightarrow\,\,$ & success   \\
        \shelf{} & reach $\rightarrow\,\,$& grasp $\rightarrow\,\,$& lift $\rightarrow\,\,$& align $\rightarrow\,\,$& approach $\rightarrow\,\,$ & & success \\
    \end{tabular}
    \vspace{-1em}
\end{table}

\subsubsection{Episode initialization and termination}

Upon initialization of each new episode, the graspable object is placed in a random location of the left bin. In tasks that contain an obstacle, the object's initial location is restricted to the regions of the space that would explicitly require the robot to circumvent the obstacle. The goal locations are initialized in the right bin, and
the robot arm is initialized at a fixed position with the gripper facing downward. Sampled initial conditions are displayed in Figure~\ref{fig:arenas}.

Each episode terminates after $H=500$ time steps. 
In addition, \push{} tasks terminate if the robot lifts the object more than a set (small) threshold above the table, in order to avoid success by the robot executing a pick-and-place strategy.

\subsection{Evaluation settings}
\label{sec:EvaluationSettings}

\benchmark{} evaluates agents for training speed and final performance over a subset of \textit{training} tasks, akin to training sets in supervised single-task settings. While this is a measure of training performance, it corresponds to the standard evaluation setting of the large majority of works in RL. After training, agents are evaluated on a \textit{test} set of unseen tasks. Both of these evaluations explore the ability of agents to discover compositional properties of the tasks.

\subsubsection{Metrics}

Agents are evaluated according to two metrics. For an agent evaluated over $N$ tasks, with $M$ evaluation trajectories for each task, each trajectory of length $H$, the average metrics are computed as follows:

\begin{minipage}[t]{0.47\textwidth}
\paragraph{Return} The standard cumulative returns:
\begin{align}
    \overline{R} = \frac{1}{NM}\sum_{i=1}^{N}\sum_{j=1}^{M}\sum_{t=1}^{H} R_i(s_t,a_t)\enspace.
\end{align}
This is the usual evaluation criterion for RL works and directly relates to the optimization objective.
\end{minipage}
\hfill
\begin{minipage}[t]{0.47\textwidth}
\paragraph{Success} The per-task success rate:
\begin{align}
    \overline{S} = \frac{1}{NM} \sum_{i=1}^{N}\sum_{j=1}^{M} \max_{t\in[1,H]}  \mathds{1} [R_{i}(s_t, a_t) = 1] \enspace,
\end{align}
where $\mathds{1}$ is the indicator function. Note that a trajectory is successful if at \textit{any} time the agent is in a success state.
\end{minipage}

\subsubsection{Evaluation on training tasks}

The agent is first evaluated on the tasks that it trains on. An agent that is capable of extracting the compositional properties of the tasks should be able to achieve transfer across the tasks. Ideally, this transfer should translate to both faster convergence in terms of the number of samples required to learn, as well as higher final performance after convergence. In particular, agents in this setting should be compared against an equivalent single-task agent that uses the same training mechanism but does so individually on every task, without any notion of shared knowledge.

\subsubsection{Evaluation on test tasks}

The key property that \benchmark{} assesses is the ability of approaches to combine trained components in novel combinations to handle new tasks. Following \citet{mendez2022modular}, this can take the following two forms:

\paragraph{Zero-shot generalization with task descriptors}
If the agent is given the multi-hot indicators described in Section~\ref{sec:observationSpace}, then it could (in principle) solve new, unseen tasks without any training on them. This would be possible only if the agent learns the compositional structure of the tasks and is able to combine its existing components into a solution to the new task. Intuitively, after learning 1)~the \pickPlace{} task with the \boxObject{} object avoiding the \objectDoor{} obstacle using the \IIWA{} arm, and 2)~the \push{} task with the \plate{} object avoiding the \objectWall{} obstacle using the \Panda{} arm, if the agent knows how each of the components relates to the overall task, it could for example swap the \IIWA{} and \Panda{} arms and solve the opposite tasks without any additional training. 

\paragraph{Few-shot generalization without task descriptors} 
Alternatively, the agent might not be informed of which components make up the current task, and be required to discover this information through experience. The goal of the agent should then be to discover this information as rapidly as possible in order to solve the new task with little experience. 

\subsubsection{Access to state decomposition}
\label{sec:StateDecomposition}
The modular architectures of \citet{devin2017learning} and \citet{mendez2022modular} require knowledge about which components of the observation affect which parts of the architecture. While this information is readily available in \benchmark{}, fair performance comparisons would require noting whether the agent is given this decomposition of the state space. Note that 
both zero-shot and few-shot settings could be targeted with or without the state decomposition.

\subsubsection{Sample of training tasks}
\label{sec:sampleTrainingTasks}

Understanding the compositional capabilities of RL algorithms requires a careful study of the sample of combinations (i.e., tasks) that is provided to the agent for training. We propose the following evaluation settings:

\paragraph{Uniform sampling} In the simplest setting, the training tasks are sampled uniformly at random, and the agent is asked to generalize to all possible combinations of the seen components. The agent therefore must learn to combine its knowledge in different ways after having seen each component in various combinations.

\paragraph{Restricted sampling} In this much harder setting, the training is restricted to a single task for one of the components and many tasks for other components (e.g., in \benchmark{}\textbackslash{}\IIWA{}, the agent sees only one \IIWA{} task and must generalize to all other \IIWA{} tasks). This is akin to Experiment~3 in the work of \citet{lake2018generalization}, which demonstrated that this is an onerous problem even in the supervised setting. While a complete evaluation would require various choices of restricted arms, objects, obstacles, and objectives, as an initial step we propose four evaluation settings: \benchmark{}\textbackslash{}\IIWA{}, \benchmark{}\textbackslash{}\hollowBox{}, \benchmark{}\textbackslash{}\objectWall{}, and \benchmark{}\textbackslash{}\pickPlace{}. These restricted elements were empirically observed to be easier to learn than others by the single-task agents during development. Restricting access to an ``easy'' element enables the zero-shot evaluation to focus on generalization, without conflating it with the difficulty of the task itself. As an exception, the \benchmark{}\textbackslash{}\objectWall{} setting was selected over \benchmark{}\textbackslash{}\noObstacle{} because generalizing to the \noObstacle{} element is trivial from any other obstacle.

\paragraph{Smaller-scale benchmarks} While large benchmarks like \benchmark{} are appealing for studying multi-task RL at scale, developing ideas in such large task sets is often (unfortunately) prohibitively time-consuming. Given the compositional nature of \benchmark{}, it is straightforward to extract smaller-scale benchmarks that maintain the properties of the full-scale benchmark. For example, \benchmark{}$\cap$\IIWA{} considers only the $64$ \IIWA{} tasks. Interestingly, such reduced benchmarks permit studying the difficulty of generalization across certain axes (e.g., if an agent can transfer knowledge across objects but not across robots, then it would perform much better on \benchmark{}$\cap$\IIWA{} than on the full \benchmark{}). Following the rationale of  the restricted setting, we propose to evaluate agents on: \benchmark{}$\cap$\IIWA{}, \benchmark{}$\cap$\hollowBox{}, \benchmark{}$\cap$\noObstacle{}, and \benchmark{}$\cap$\pickPlace{}.

\section{Benchmarking existing RL methods on \benchmark{}} \label{sec:experiments}

The empirical evaluation in this section had two primary objectives. First, to demonstrate that \benchmark{} is a useful evaluation benchmark in terms of: 1)~existing algorithms making progress toward solving the problems, 2)~the tasks exhibiting compositional properties, and 3)~existing approaches leaving substantial room for improvement in performance. 
Second, to provide benchmarking results of existing algorithms for future work to leverage
\footnote{Trained models are available at: \url{https://github.com/Lifelong-ML/CompoSuite-Data}.}.

\subsection{Experimental setting}

The underlying RL algorithm used for all our evaluations was the proximal policy optimization (PPO; \citealp{schulman2017proximal}) implementation in \texttt{Spinning\! Up}~\citep{SpinningUp2018}. Appendix~\ref{app:PPODetails} describes critical modifications that were necessary for learning the tasks. Building upon this base algorithm, we evaluated the following three agents:
\vspace{-0.5em}
\begin{itemize}[leftmargin=*,noitemsep,topsep=0pt,parsep=0pt,partopsep=0pt]
    \item \textbf{Single-task} agents that trained on each task individually, without any knowledge-sharing across tasks. Lack of sharing precludes these agents from generalizing to unseen tasks, and so they were only evaluated on training tasks. We also withheld the task descriptor from the observation, as it would appear as a constant to each single-task agent.
    \item \textbf{Multi-task} agents that trained a {\em shared model} for all tasks, using the task descriptor in the observation to help differentiate between tasks and learn to specialize the policy for each task. Given the need for the multi-task agent to encode multiple policies in a single model, we gave this agent a larger capacity than an individual single-task agent.
    \item \textbf{Compositional} agents that constructed a \textit{different} model for each task from a set of \textit{shared} components. We used a variant of the modular network of \citet{mendez2022modular} that establishes each policy from a set of modules, with one module for each robot, object, obstacle, and objective. The relevant state component from Section~\ref{sec:observationSpace} was fed as input to each module, and the task descriptor was used to select the correct modules. Each module was represented by an MLP, whose output was fed to the next module: \textsf{obstacle} $\rightarrow$ \textsf{object} $\rightarrow$ \textsf{objective} $\rightarrow$ \textsf{robot} (see Appendix~\ref{app:PPODetails}). For fairness, the \textit{overall} number of parameters across modules was equivalent to that of multi-task agents. 
\end{itemize}

Each agent was evaluated on the full \benchmark{} benchmark and on all the suggested smaller-scale and restricted benchmarks. In each of these settings, a subset of the tasks was given to the agents for training, and the agents were evaluated for their speed and final performance over the training tasks. After training, the multi-task and compositional agents were additionally evaluated for their ability to solve unseen tasks without any additional training by leveraging the task descriptors. Additional details are provided in Appendix~\ref{app:PPODetails}. 

\subsection{Evaluation of baselines on the full \benchmark{} benchmark}
\label{sec:MTexp}

\begin{figure}[t!]
\captionsetup[subfigure]{font=scriptsize, aboveskip=2pt}
\centering
    \begin{subfigure}[b]{0.7\textwidth}
            \centering
                \includegraphics[width=\linewidth]{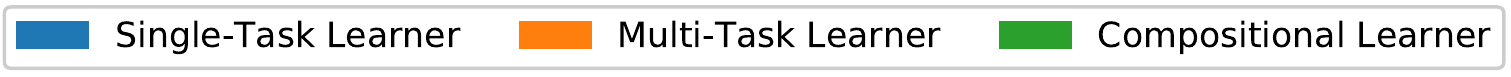}
        \end{subfigure}\\
    \begin{subfigure}[b]{0.28\textwidth}
    \centering
        \includegraphics[height=3cm, trim={0.2cm 0.0cm 1.5cm 0.8cm}, clip]{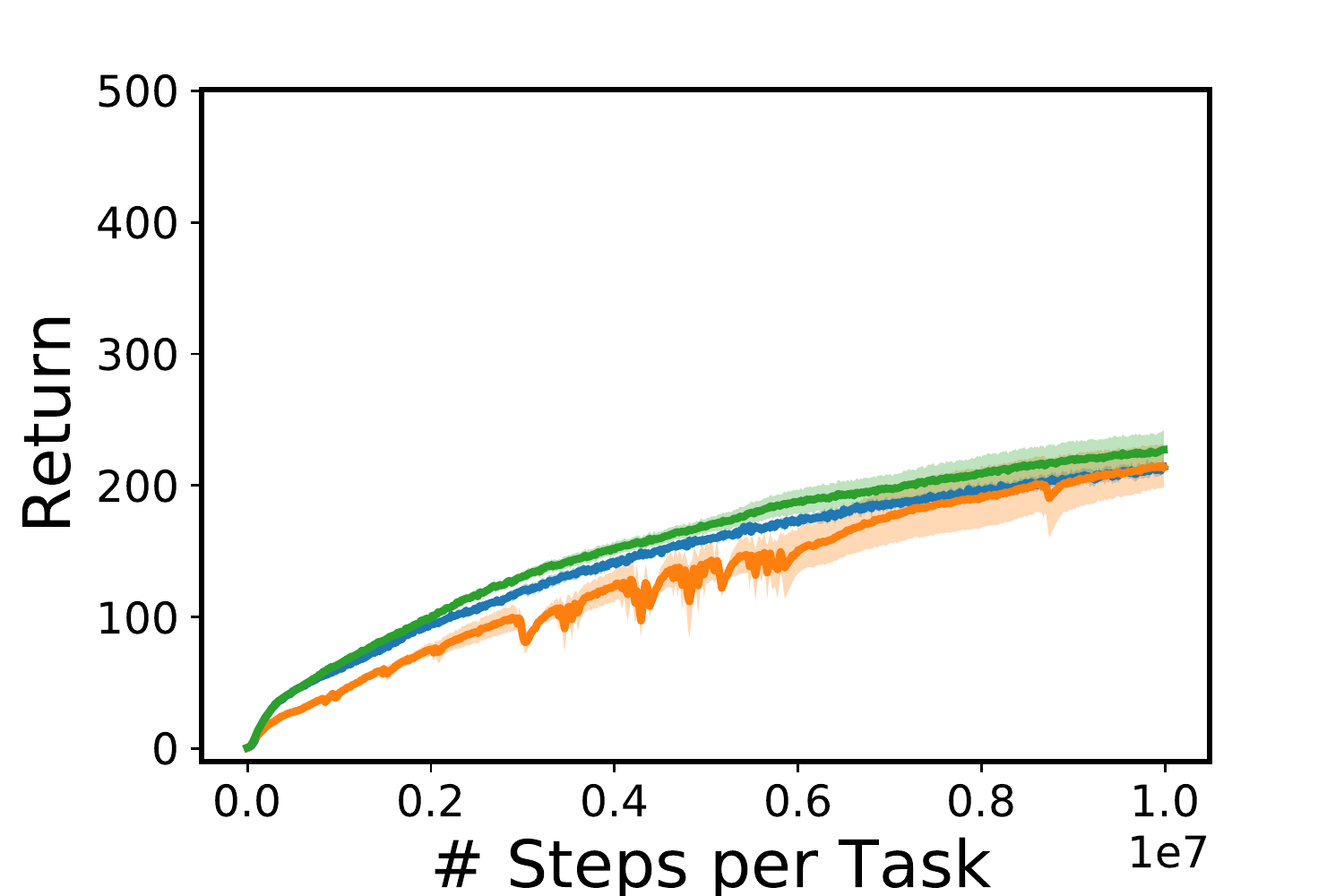}\\
        \vspace{-8.5em}
        {\hspace{20pt}$56$ training tasks}
    \end{subfigure}%
    \begin{subfigure}[b]{0.24\textwidth}
    \centering
        \includegraphics[height=3cm, trim={2.2cm 0.0cm 1.5cm 0.8cm}, clip]{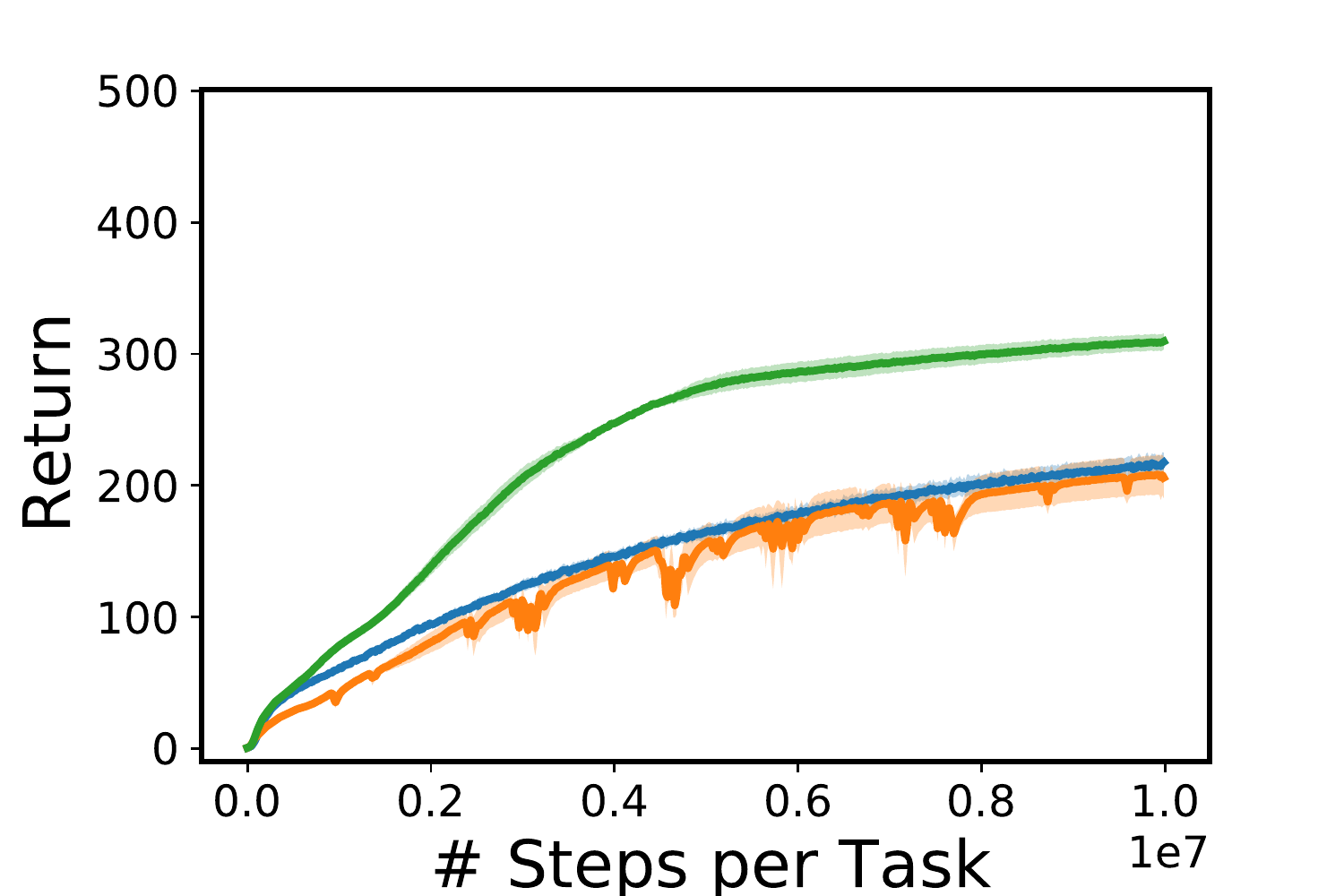}\\
        \vspace{-8.5em}
        {$224$ training tasks}
    \end{subfigure}%
    \begin{subfigure}[b]{0.24\textwidth}
    \centering
        \includegraphics[height=3cm, trim={2.2cm 0.0cm 1.5cm 0.8cm}, clip]{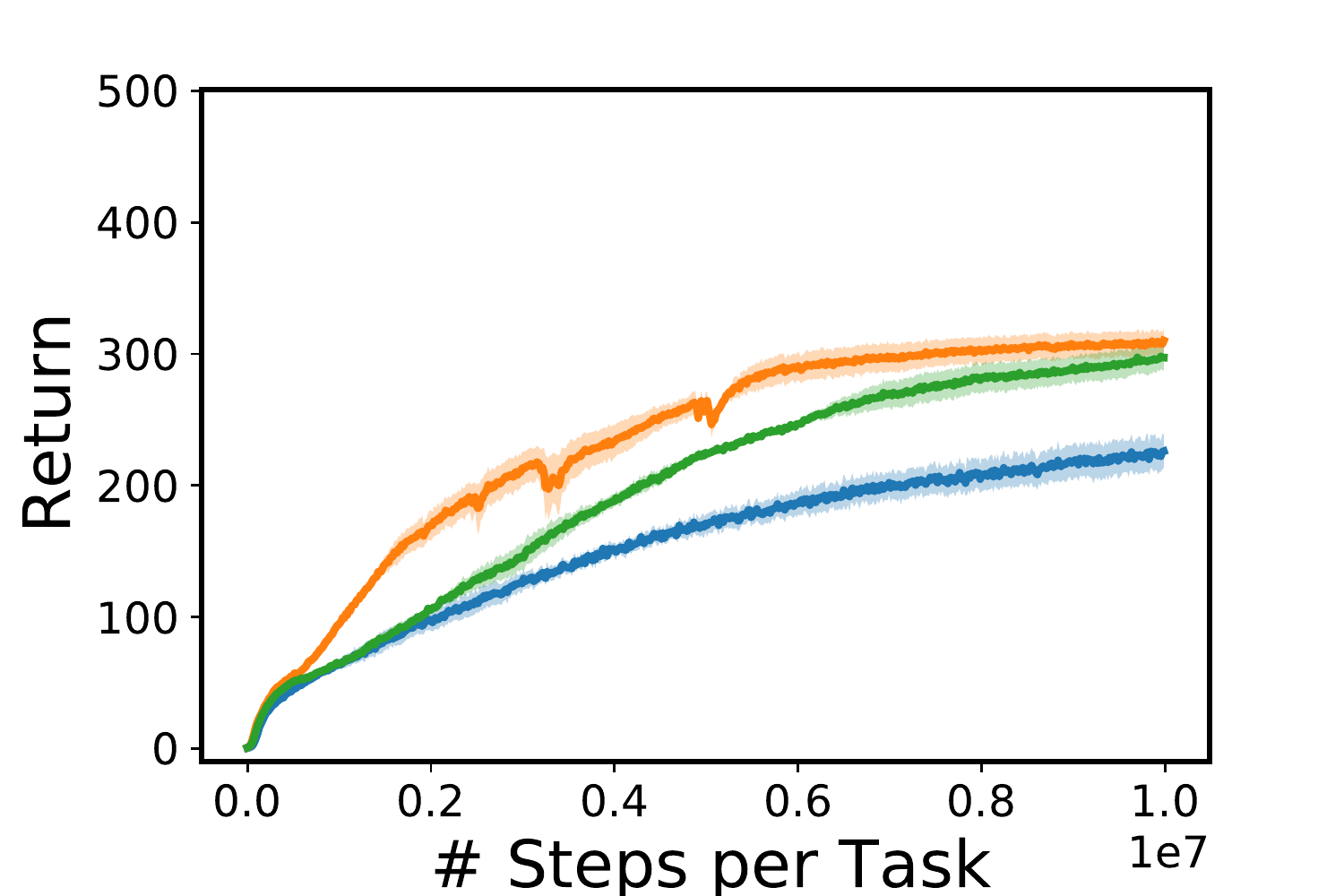}\\
        \vspace{-8.5em}
        {$32$ training tasks}
    \end{subfigure}%
    \begin{subfigure}[b]{0.24\textwidth}
    \centering
        \includegraphics[height=3cm, trim={2.2cm 0.0cm 1.5cm 0.8cm}, clip]{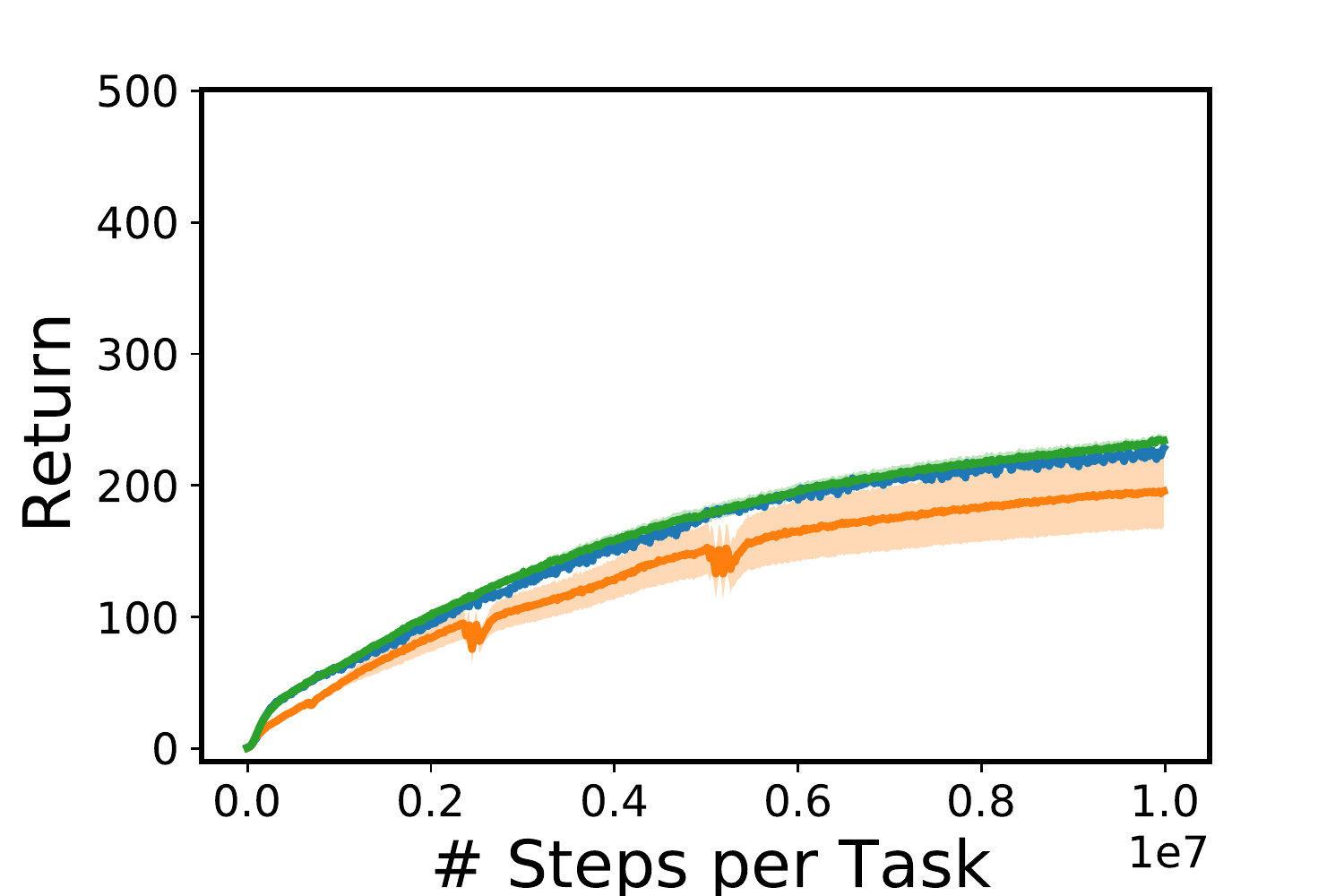}\\
        \vspace{-8.5em}
        {$56$ training tasks}
    \end{subfigure}%
    \vspace{7.1em}
    \\
    \begin{subfigure}[b]{0.28\textwidth}
    \centering
        \includegraphics[height=3cm, trim={0.2cm 0.0cm 1.5cm 0.8cm}, clip]{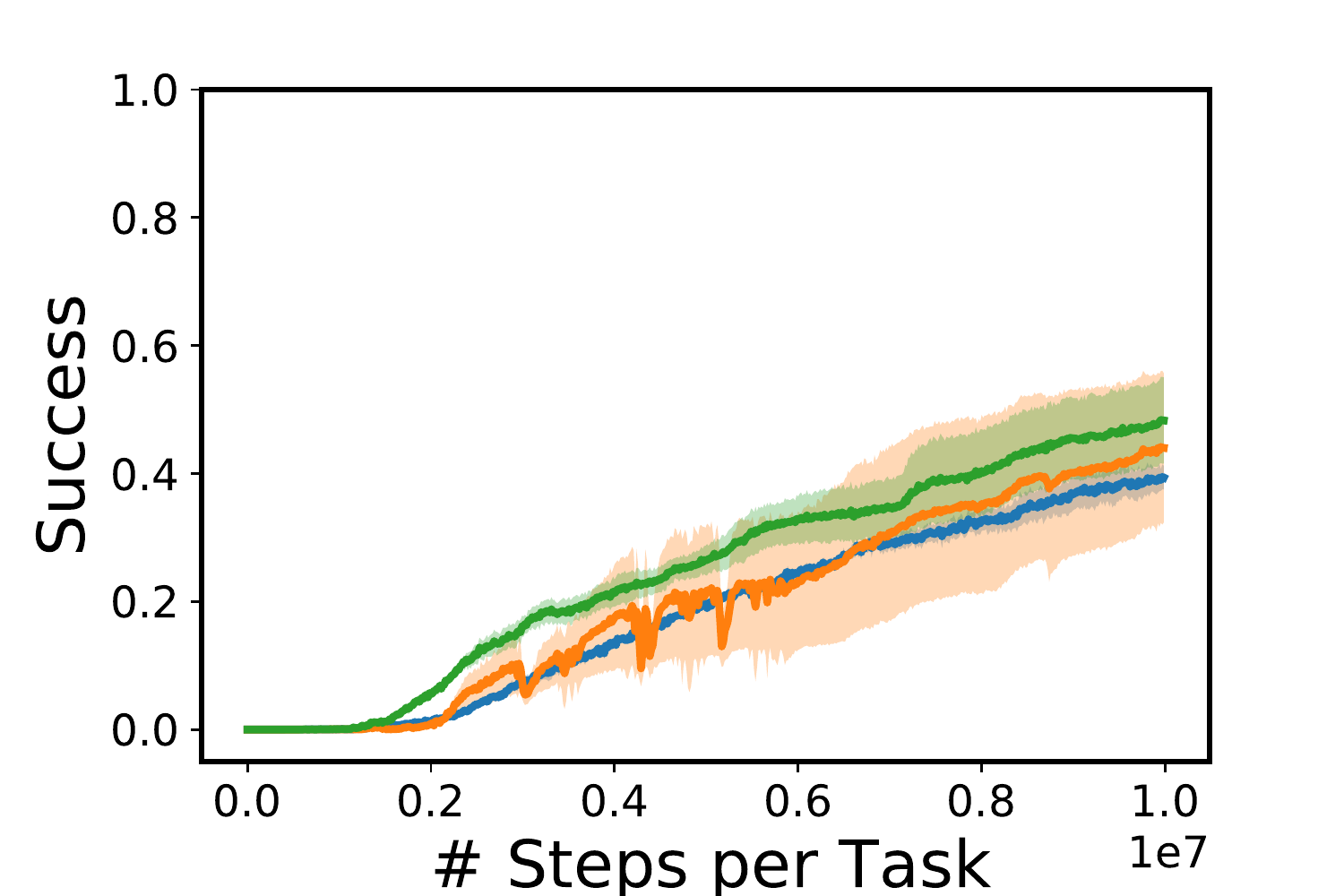}
        \caption{Full \benchmark{}}
        \label{fig:fullBenchmarkCurvesSmall}
    \end{subfigure}%
    \begin{subfigure}[b]{0.24\textwidth}
    \centering
        \includegraphics[height=3cm, trim={2.2cm 0.0cm 1.5cm 0.8cm}, clip]{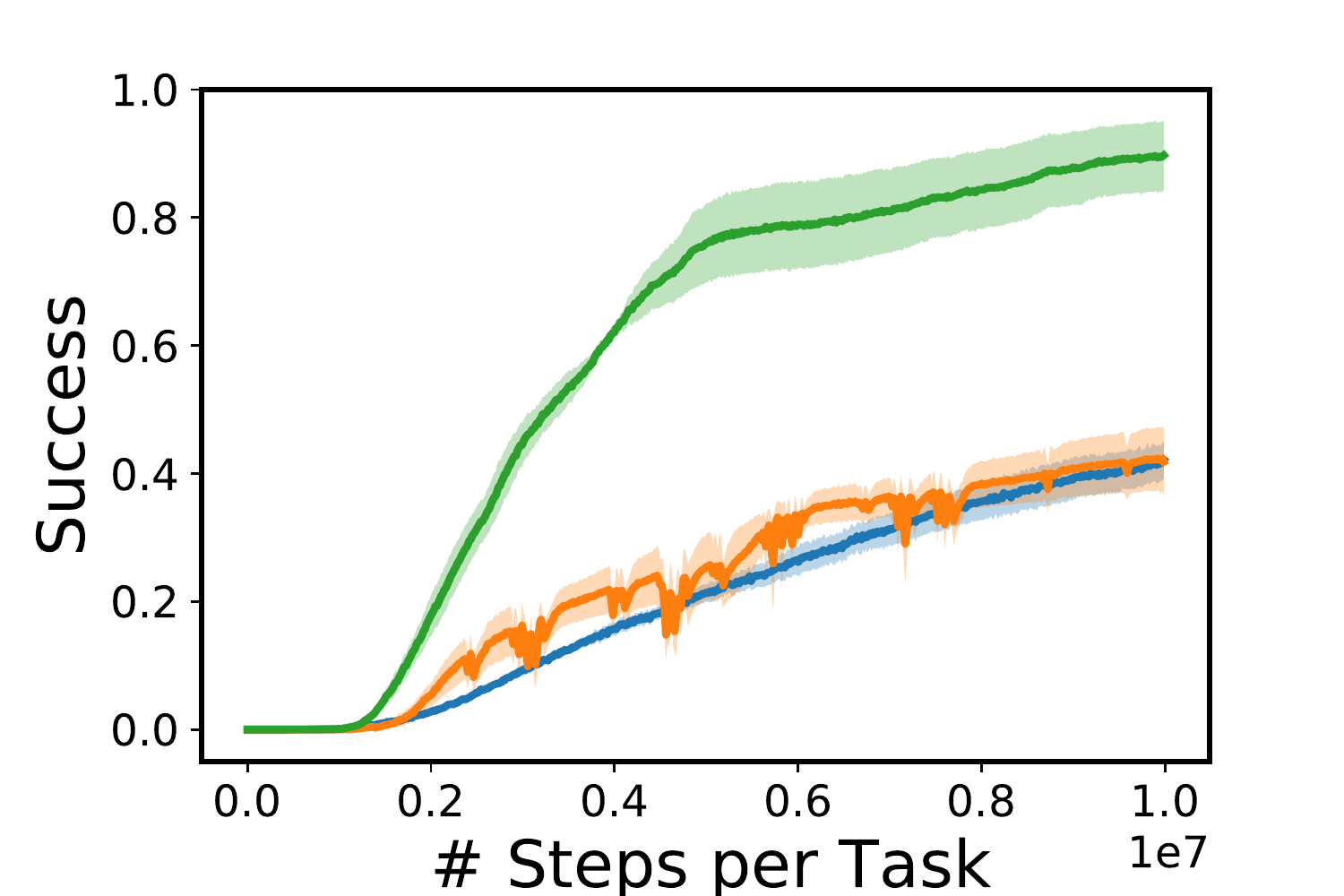}
        \caption{Full \benchmark{}}
        \label{fig:fullBenchmarkCurvesLarge}
    \end{subfigure}%
    \begin{subfigure}[b]{0.24\textwidth}
    \centering
        \includegraphics[height=3cm, trim={2.2cm 0.0cm 1.5cm 0.8cm}, clip]{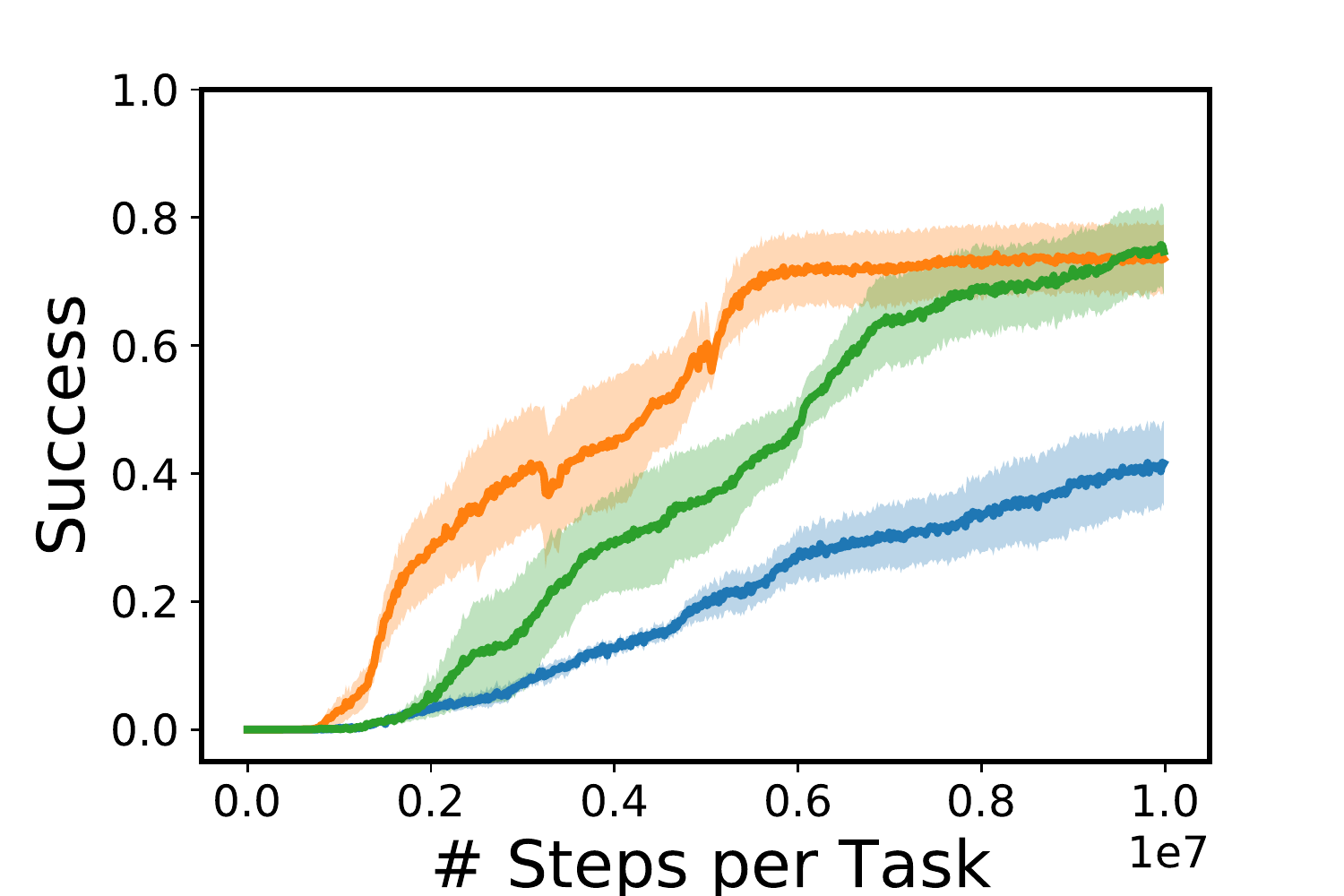}
        \caption{\benchmark{}$\cap$\IIWA{}}
        \label{fig:smallscaleCurves}
    \end{subfigure}%
    \begin{subfigure}[b]{0.24\textwidth}
    \centering
        \includegraphics[height=3cm, trim={2.2cm 0.0cm 1.5cm 0.8cm}, clip]{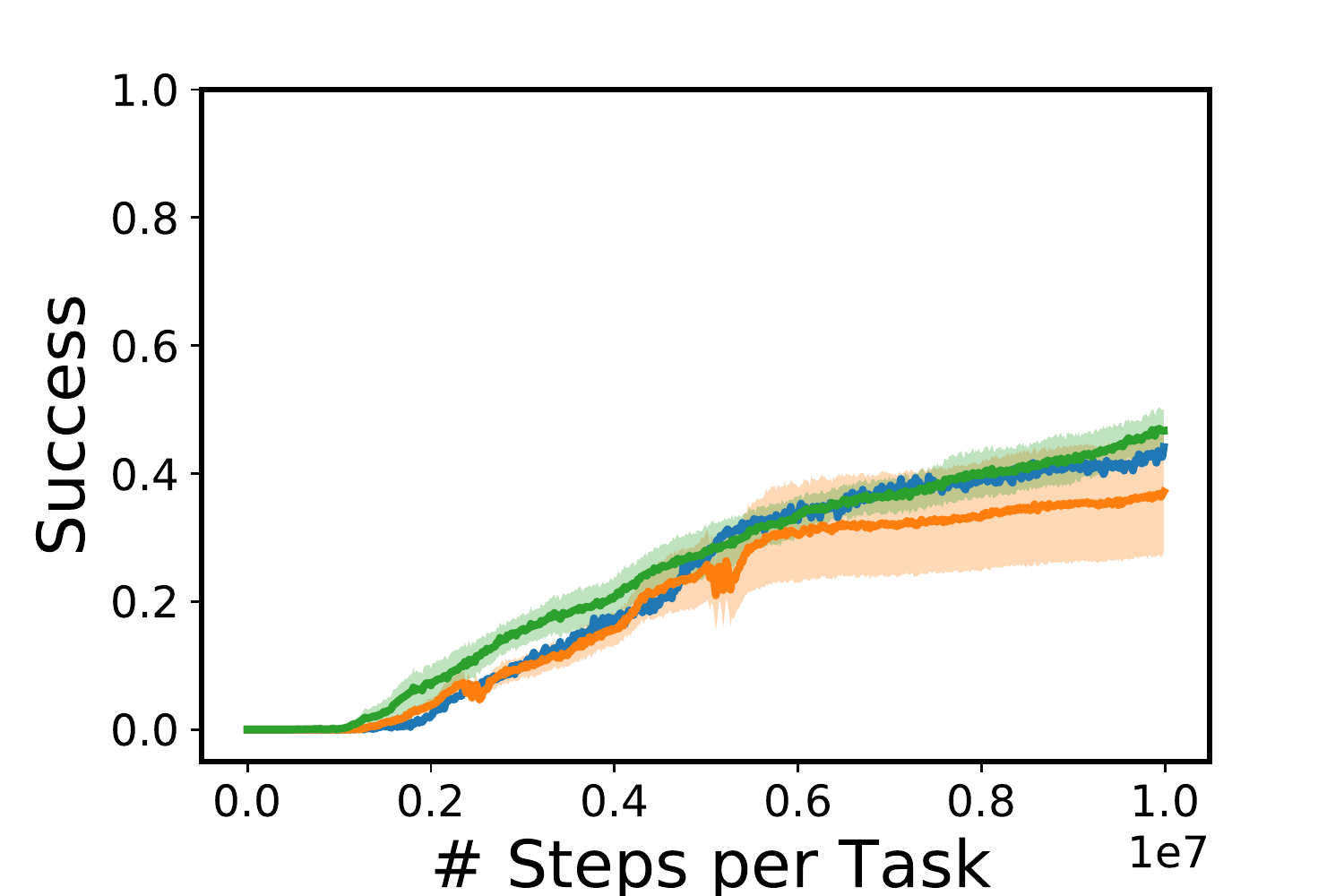}
        \caption{\benchmark{}\textbackslash{}\pickPlace{}}
        \label{fig:holdoutCurves}
    \end{subfigure}%
    \vspace{-0.5em}
    \caption{Evaluation on training tasks. (a,b) When trained on all of \benchmark{}, the multi-task agent was not capable of accelerating the learning substantially with respect to the single-task agent according to either metric. (b) The compositional agent, when trained on a larger set of tasks, performed noticeably better, demonstrating that leveraging the compositional structure of \benchmark{} leads to improved training performance. (c) On the smaller-scale  \benchmark{}$\cap$\IIWA{}, both the multi-task and the compositional agents were able to outperform the single-task agent under both metrics. This shows that sharing knowledge across tasks with a single robot arm is easier than across different robot arms when not explicitly leveraging compositionality. (d) On \benchmark{}\textbackslash{}\pickPlace{}, a single \pickPlace{} task was sampled for training, and all other tasks were sampled from the remaining task objectives. Results are qualitatively similar to those on the full \benchmark{}, as expected---major differences are expected in \textit{zero-shot} performance. Y-axes span the attainable ranges and shaded regions represent std.~errors across three seeds.}
    \label{fig:learningCurves}
\end{figure}
We first evaluated the agents on the main \benchmark{} benchmark, uniformly sampling tasks for training; learning curves are presented in Figures~\ref{fig:fullBenchmarkCurvesSmall} and~\ref{fig:fullBenchmarkCurvesLarge}. After training for $10$ million time steps on each task, the single-task agent had a success rate of around $40\%$. When the training set was a small portion of the whole set of tasks, the multi-task and compositional agents only slightly improved upon the single-task agent. However, when training on a larger set of tasks, the compositional agent learned much faster, achieving approximately twice as much success. In contrast, multi-task results did not improve with the larger training set. This suggests that the multi-task agent is not appropriately sharing knowledge across tasks, and instead separately allocates capacity in the network to different tasks. As more tasks are seen, capacity is progressively exhausted. Instead, the compositional agent shares components appropriately, and additional training tasks improve the agent's ability to leverage these commonalities. This demonstrates that \benchmark{} tasks are indeed compositionally related, and that exploiting these relationships leads to improved performance.

After training, we evaluated the agents on the \benchmark{} tasks that they did \textit{not} train on. Intuitively, an agent that correctly decomposes the tasks should achieve high performance on these test tasks by adequately recombining its learned components. Results in the first two columns of Table~\ref{tab:zeroshot} show that the learners struggled to generalize to unseen tasks when trained on $56$ tasks but performed remarkably well when trained on $224$ tasks. With the smaller training set, even though the \textit{training} performance was similar for both approaches, the compositional agent achieved substantially worse \textit{zero-shot} performance. This demonstrates that, while the compositional approach can indeed capture the compositional properties of the tasks, this capability requires observing a large portion of the tasks.

One important question is whether the multi-task agent was automatically learning compositional knowledge that allowed it to solve unseen tasks. The alternative explanation would be that the agent instead found similar tasks in the training set and used the policy for those for generalization. We therefore set up a simple experiment, finding the most similar training task to each test task and using its policy to predict zero-shot performance. Concretely, for every test task $\mathcal{M}_i$ with some zero-shot success, we found the training task $\mathcal{M}_{i'}$ whose policy $\pi_{i'}$ performed best on $\mathcal{M}_i$; this would have been the best policy to choose, and so one would expect the performance of $\pi_{i'}$ to correlate to that of $\pi_{i}$. 
To reduce computation, we considered only tasks $i'$ that varied in a single element from $i$. We found that the coefficients of determination between the policies' success rates were very low: $\mathbf{R^2=0.19}$ and $\mathbf{R^2=0.03}$ for the multi-task and compositional agents, respectively. This shows that the generalization of the multi-task learner was unlikely to come from using trained policies for different tasks, but rather from leveraging the compositional properties of the tasks.

\begin{table}[t!]
\addtolength{\tabcolsep}{-0.1em}
    \captionof{table}{\benchmark{} zero-shot generalization. All agents struggled to generalize to the majority of the holdout tasks except for the compositional agent trained on $224$ tasks; std.~errors across three seeds reported after the $\pm$.}
    \label{tab:zeroshot}
    \centering
    \vspace{-1em}
        \begin{tabular}{l|ll|ll|ll|ll}
         & \multicolumn{2}{c|}{\benchmark{}} & 
         \multicolumn{2}{c|}{\benchmark{}} & 
         \multicolumn{2}{c|}{\benchmark{}} & \multicolumn{2}{c}{\benchmark{}}\\
         & \multicolumn{2}{c|}{$56$ Tasks} & 
         \multicolumn{2}{c|}{$224$ Tasks} & 
         \multicolumn{2}{c|}{$\cap$\IIWA{}} & \multicolumn{2}{c}{\textbackslash{}\pickPlace{}}\\
         \hline\hline
         & Multi-task & Comp. & Multi-task & Comp. & Multi-task & Comp. & Multi-task & Comp.\\
         \hline
        Return & $115.79${\tiny$ \pm 18.0$} & $64.26${\tiny$\pm 7.5$} & $201.74${\tiny$\pm 26.9$} & $302.44${\tiny$\pm 12.2$} & $232.79${\tiny$\pm 17.5$} & $79.85${\tiny$\pm 19.2$} & $74.61${\tiny$\pm 10.$} & $16.63${\tiny$\pm 6.7$}\\
        Success & $0.18${\tiny$ \pm 0.1$} & $0.08${\tiny$\pm 0.0$} & $0.41${\tiny$\pm 0.0$} & $0.88${\tiny$\pm 0.1$} & $0.49${\tiny$\pm 0.1$} & $0.12${\tiny$\pm 0.1$} & $0.09${\tiny$\pm 0.0$} & $0.01${\tiny$\pm 0.0$}\\
        \end{tabular}
    \vspace{-1em}
\end{table}

\subsection{Evaluation of baselines on the smaller-scale \benchmark{}\texorpdfstring{$\cap$}{∩}\IIWA{} benchmark} 
\label{sec:smallscaleExperiments}

Next, we evaluated the three baseline agents on the reduced benchmarks, in order to 1)~propose a computationally cheaper setting to facilitate progress and 2)~shed light on the relative difficulty of generalizing across different \benchmark{} axes. Learning curves on the training tasks for the \benchmark{}$\cap$\IIWA{} benchmark are included in Figure~\ref{fig:smallscaleCurves}, and the remaining curves are in Appendix~\ref{app:additionalSampling}. The relative performance of the compositional agent with respect to the single-task agent was close to that obtained after training on \textit{over $\mathit{200}$ tasks} for the full \benchmark{}, demonstrating that the agent is capable of discovering the compositional structure of this reduced benchmark with far fewer training tasks. On the other hand, the multi-task agent performed noticeably better in \benchmark{}$\cap$\IIWA{} and \benchmark{}$\cap$\pickPlace{}. This provides evidence that these two axes (robot and objective) are harder to generalize across, as one would intuitively expect. Additional evidence toward this hypothesis is given in Section~\ref{sec:additionalExperiments}.

We further assessed the performance of the agents on the unseen \IIWA{} tasks, and report the results in the third column of Table~\ref{tab:zeroshot}. The multi-task agent achieved notably high performance but the compositional agent was incapable of generalizing. The poor generalization of the compositional agent was likely due to the small number of training tasks: since the agent only trains each module on the subset of tasks that shares that module, each parameter was trained on a small number of tasks which was insufficient for zero-shot generalization. Results on the remaining settings in Appendix~\ref{app:additionalSampling} exhibit lower multi-task performance and similar compositional performance in most cases.

\subsection{Evaluation of baselines on the restricted \benchmark{}\textbackslash{}\pickPlace{} benchmark} \label{sec:holdoutExperiments}

\begin{figure}[t]
\begin{minipage}[t]{.29\linewidth}
\addtolength{\tabcolsep}{-0.25em}
\captionsetup{font=small}
\vspace{-1.8em}
    \centering
    \captionof{table}{Zero-shot generalization (success rate) for the multi-task agent on \benchmark{}\textbackslash{}\breakablePickPlace{}, separated by tasks that share (or not) each element with the trained \breakablePickPlace{} task (e.g., if the training \breakablePickPlace{} task used the \IIWA{} arm, \IIWA{} tasks go in the ''trained'' column and non-\IIWA{} tasks go in the ``untrained'' column).}
    \label{tab:deaggreagtedZeroshotHoldout}
    \vspace{-.3em}
    \begin{tabular}{@{}l|c|c@{}}
        Element & Trained & Untrained \\
         \hline
        robot    & $0.30${\tiny$ \pm 0.10$} & $0.03${\tiny$ \pm 0.02$} \\
        object   & $0.14${\tiny$ \pm 0.04$} & $0.08${\tiny$ \pm 0.04$} \\
        obstacle & $0.14${\tiny$ \pm 0.04$} & $0.08${\tiny$ \pm 0.03$} \\
    \end{tabular}
\end{minipage}
\hfill
\begin{minipage}[t]{0.69\linewidth}
\captionsetup[subfigure]{aboveskip=1pt}
\centering
    \begin{subfigure}[b]{\textwidth}
        \centering
            \includegraphics[width=0.8\linewidth]{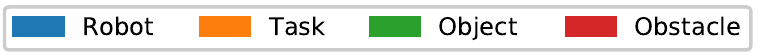}
    \end{subfigure}\\
    \vspace{.2em}
    \begin{subfigure}[b]{0.26\textwidth}
        \centering
            \includegraphics[height=2.4cm, trim={0.0cm, 0.15cm, 0.35cm, 0.6cm}, clip]{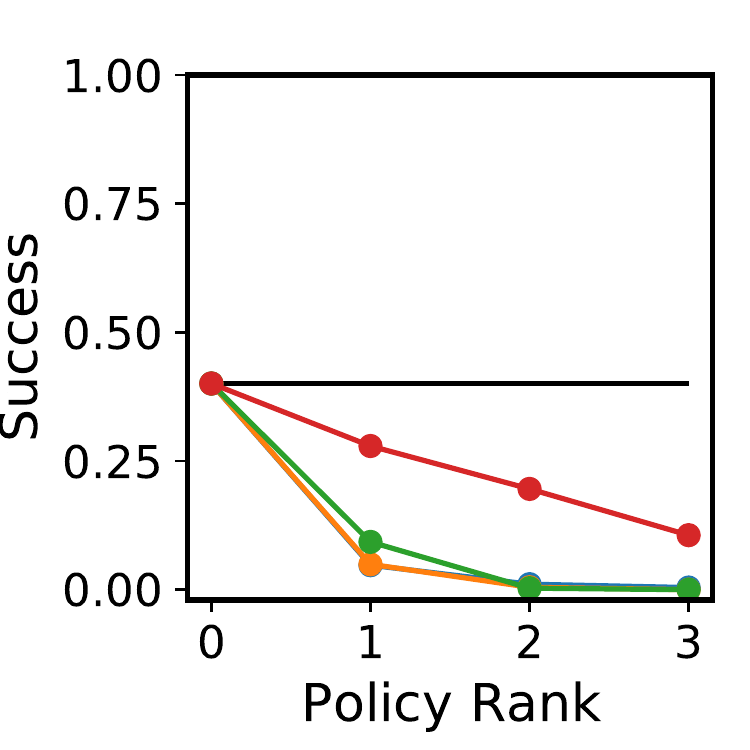}
        \caption*{MT, train}
    \end{subfigure}%
    \begin{subfigure}[b]{0.248\textwidth}
        \centering
            \includegraphics[height=2.4cm, trim={0.5cm, 0.15cm, 0.35cm, 0.6cm}, clip]{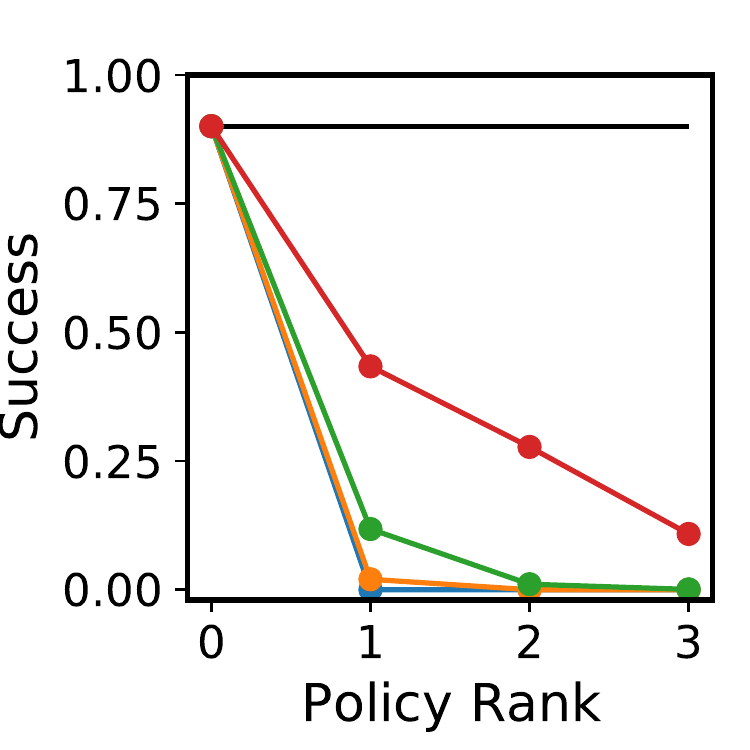}
        \caption*{Comp., train}
    \end{subfigure}%
    \begin{subfigure}[b]{0.248\textwidth}
        \centering
            \includegraphics[height=2.4cm, trim={0.5cm, 0.15cm, 0.35cm, 0.6cm}, clip]{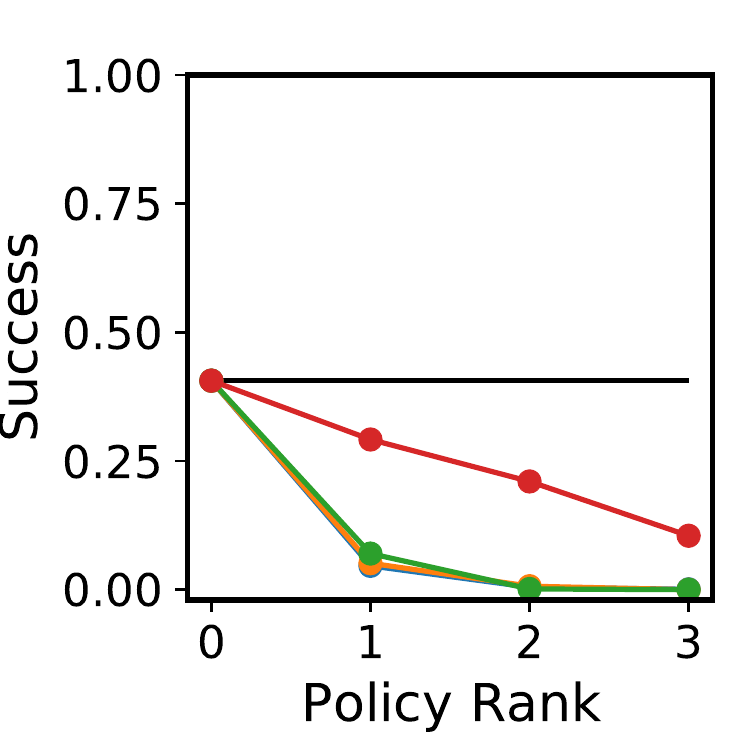}
        \caption*{MT, zero-shot}
    \end{subfigure}%
    \begin{subfigure}[b]{0.248\textwidth}
        \centering
            \includegraphics[height=2.4cm, trim={0.5cm, 0.15cm, 0.35cm, 0.6cm}, clip]{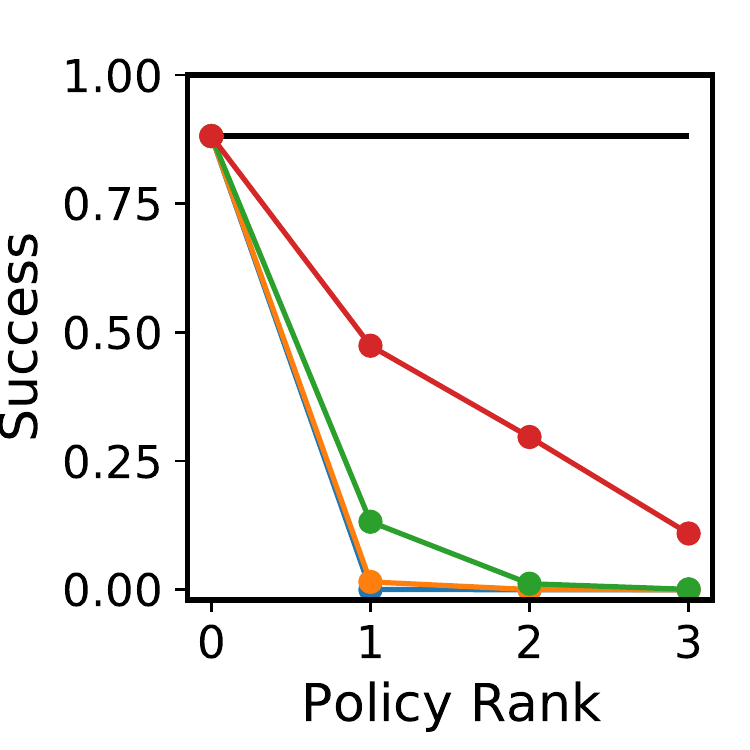}
        \caption*{Comp., zero-shot}
    \end{subfigure}%
    \vspace{-0.3em}
    \caption{Performance given incorrect task descriptors, with a single changed component: position $0$ corresponds to the correct descriptor, and positions $j>0$ correspond to the $j$-th ranked descriptors of each type. Using the wrong descriptors leads to poor performance, confirming the diversity of \benchmark{} tasks.}
    \label{fig:task_sim_mtl}
\end{minipage}
\end{figure}

The results presented so far consider a relatively simple compositional problem: the agent is trained on multiple combinations of all components, and is expected to generalize to new combinations. These previous results already expose the shortcomings of existing approaches in the compositional setting. However, future approaches that solve these simple settings would still fall short from achieving the full compositional capabilities we expect from them. 
We would hope that agents could learn components that generalize to unseen tasks even if these components are only seen in one single combination. To study this setting, we evaluated the three agents on the restricted task samples. Results on the training tasks for the  \benchmark{}\textbackslash{}\pickPlace{} benchmark, which includes exactly one \pickPlace{} task, are shown in Figure~\ref{fig:holdoutCurves}, and Appendix~\ref{app:additionalSampling} includes
results for remaining settings. In many cases, performance was close to that of the full benchmark using $56$ tasks, 
because the training distributions are 
similar: there are $55$ combinations 
of $15$ components (plus one restricted task), compared to $56$ combinations of $16$ components.
The fourth column of Table~\ref{tab:zeroshot} summarizes the \pickPlace{} zero-shot performance of the agents on this restricted setting. Both agents failed to generalize to the unseen \pickPlace{} tasks (Appendix~\ref{app:additionalSampling} shows the same trend for the remaining settings). Table~\ref{tab:deaggreagtedZeroshotHoldout} shows that the small amount of generalization the multi-task agent achieved was almost entirely on tasks from the same robot arm that was used in the single \pickPlace{} training task. 

\subsection{Empirical analysis of \benchmark{} learnability and diversity}
\label{sec:additionalExperiments}

We now shift focus to verifying two important properties of \benchmark{}: that the large majority of tasks are learnable by current RL mechanisms, and that the tasks are not only compositional, but also highly varied.

\paragraph{Learnability of tasks} If \benchmark{} tasks were unsolvable by current RL methods, that would conflate the difficulty of compositional reasoning with the difficulty of solving RL tasks. To validate that this is not the case, for every task, we found the best performing agent across all those trained in Figure~\ref{fig:learningCurves} (taking the maximum across experiments \textit{and} random seeds). For any task with a score of $0$, we have no evidence that the task is learnable, because no agents solved it to any extent. The result of this computation is shown for all tasks in  Appendix~\ref{app:MaxSuccessRatePerTask}. Only one task received a score of $0$, indicating that it {\em may} be unlearnable; this demonstrates that \benchmark{} tasks are attainable for current RL methods. 

\textbf{Diversity of tasks}~~Another valid concern is that it might be possible for the agent to solve multiple tasks with a single policy if the tasks are very similar, implying that compositional reasoning is not necessary for generalization. To verify that this is not the case in \benchmark{}, after training the multi-task and compositional agents over $224$ tasks, we evaluated their performance if they were given the \textit{incorrect} task descriptor. In particular, for a given task $\mathcal{M}_i$, we tested the performance of the agent on task $\mathcal{M}_i$ if given the descriptor for all tasks $\mathcal{M}_{i'}$ that varied in a single component from $\mathcal{M}_i$ (i.e., the tasks most similar to $i$). We sorted the performances with these incorrect descriptors separately for each axis, finding the rank of each incorrect component (e.g., the rank-two robot for task $\mathcal{M}_i$ is the robot that achieved the second-best performance when
used as the descriptor for task $\mathcal{M}_i$), and averaged the sorted performances. 
The results, summarized in Figure~\ref{fig:task_sim_mtl}, show 
that using the
incorrect robot, objective or object
descriptor consistently leads to substantially degraded
performance, particularly for the compositional agent; this means that the agents are specializing to each of the components. Using the incorrect obstacle descriptor has a much smaller impact, particularly for the multi-task agent, which suggests that the multi-task agent learns a policy that is somewhat agnostic to the obstacles. We conclude that \benchmark{} cannot be solved without specializing the policies to each task. Additionally, varying the robot arm causes a drastic drop in performance, demonstrating that solving tasks with varied robots is a challenging problem, yet existing benchmarks are limited to a single robot arm.

\section{Scope, limitations, and extensions}

\benchmark{} is designed as a benchmark for studying the compositional properties of multi-task RL algorithms. As such, while it can be used to investigate multiple other problems, it is not intended to cover the spectrum of open questions in multi-task RL. This section discusses limitations and potential extensions to the use of \benchmark{}.

\paragraph{Reliance on PPO} To provide a fair comparison across single-task and multi-task learners, we used PPO~(\citealp{schulman2017proximal}) as the base RL algorithm for all agents, built off of the \texttt{Spinning\! Up} implementation~\citep{SpinningUp2018}. While future research can use any base learning method, only evaluations that use the same PPO implementation could fairly compare against the benchmarking results presented here.

\paragraph{Input space} The input space used in our evaluation is a $94$-dimensional symbolic description of the environment grounded in the system dynamics. However, there is also broad interest from the robot learning community in RL with richer observations (e.g., visual inputs). While such an evaluation falls outside of the scope of this work, the benchmark implementation allows users to request a multi-camera visual observation instead of the low-dimensional observation.

\paragraph{Task descriptors} Part of the observation space is a multi-hot indicator that describes the components that make up the current task. While this permits assessing the interesting property of zero-shot compositional generalization, there are other questions that might benefit from withholding this information from the agent. As one example, the agent might be given only a task index that indicates \textit{which} task is currently being solved, but not \textit{how} it relates to other tasks. Alternatively, the agent could be given no indication of the current task at all and be required to extract it from data. Note that the symbolic observation does not contain sufficient information to unequivocally identify the task without task descriptors, and so the agent would need to extract this information from \textit{trajectories} of interaction instead. On the other hand, the images in the visual observations do contain sufficient information to differentiate the tasks.

\paragraph{Additional compositional axes} \benchmark{} currently consists of four compositional axes, in part to constrain the size of the benchmark---the number of tasks is exponential in the number of axes. Expanding the benchmark with new axes could be necessary as deep RL methods scale further. As examples, new axes could include discretized object placements at initialization, goal positions, or variations in textures and color if using visual inputs.

\paragraph{Other forms of composition} While \benchmark{} was designed around functional composition as described in Section~\ref{sec:problem}, the benchmark can also be used for other forms of composition. In particular, the standardization of the environments and the use of stage-wise rewards makes this a useful domain for evaluating skill discovery and sequencing. For example, the agent could learn skills for reaching a location, grasping an object, and lifting, all of which are useful for multiple \benchmark{} tasks. Note that standard representations of skills would only work for one individual arm.

\paragraph{Continual learning} Another very natural extension of \benchmark{} is to use it in the continual learning setting, particularly when viewed as an online version of multi-task learning. The agent would be presented with \benchmark{} tasks one after the next, and evaluated on all previously seen tasks. The goal of the agent would be to learn each new task as quickly as possible by leveraging accumulated knowledge, and to retain performance on the earlier tasks upon training on new tasks. Given the sequential nature of continual learning, it might be prohibitively expensive to train the agent over the full variant of \benchmark{}, but the smaller-scale variants described in Section~\ref{sec:sampleTrainingTasks} would be feasible; existing approaches have already been evaluated on similar-length sequences of robotics tasks~\citep{mendez2022modular}. 

\paragraph{Sim2real transfer} Learning a multitude of tasks in simulation is a common strategy used to transfer policies from simulation to the real world (sim2real). Since \benchmark{} uses simulated versions of four robot arms that are commercially available, it could additionally be leveraged to study this promising direction.

\section{Conclusions} 
\label{sec:conclusion}

We introduced \benchmark{}, a large-scale robotic manipulation benchmark for studying the novel problem of functionally compositional RL. \benchmark{} leverages the power of combinatorics to create hundreds of highly diverse tasks, opening the door for multi-task RL at scale. In particular, \benchmark{} is designed to study the ability of approaches to discover the decomposition of complex problems into simpler subproblems whose solutions can be combined to solve the overall task. Once appropriate components have been found, they could be combined to solve \textit{new} RL problems that the agent has never trained on. Existing end-to-end and modular multi-task approaches show promising properties in some limited settings under \benchmark{}, but we expose that they are far from solving the problem of compositional RL. Progress in that direction will enable RL approaches to automatically detect commonalities across diverse problems, leverage these commonalities to facilitate learning, and eventually handle far more complex tasks than is possible today.

\subsubsection*{Acknowledgments}
We would like to thank Spencer Solit for help in the early stages of development of \benchmark{}, and David Kent for technical feedback about the robot simulator. We also thank the anonymous reviewers for their valuable comments. The research presented in this paper was partially supported by the DARPA Lifelong Learning Machines program under grant FA8750-18-2-0117, the DARPA SAIL-ON program under contract HR001120C0040, the DARPA ShELL program under agreement HR00112190133, and the Army Research Office under MURI grant W911NF20-1-0080.

\setcitestyle{numbers}
\bibliography{CompoSuite}
\bibliographystyle{collas2022_conference}
\setcitestyle{authoryear}
\newpage

\appendix

\section{Visualization of all tasks}
\label{app:VisualizationOfAllTasks}

\begin{figure}[H]
    \vspace{2.25em}
        \includegraphics[width=0.98\linewidth, trim={0.3cm 0.0cm 1.5cm 0.8cm}]{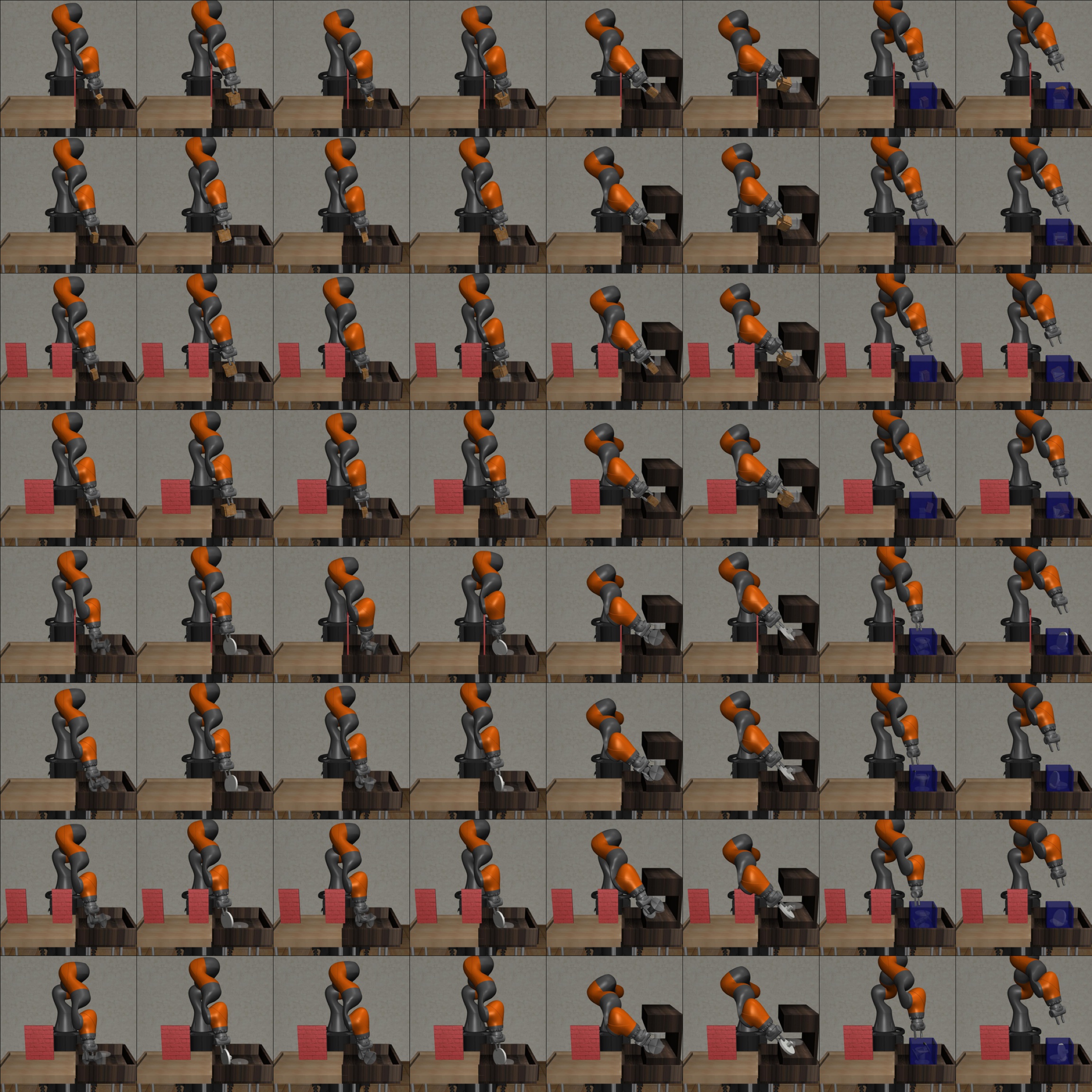}
    \caption{Visualization of the $64$ \IIWA{} tasks.}
    \label{fig:iiwa_viz}
\end{figure}

\begin{figure}[p]
        \includegraphics[width=0.98\linewidth, trim={0.3cm 0.0cm 1.5cm 0.8cm}]{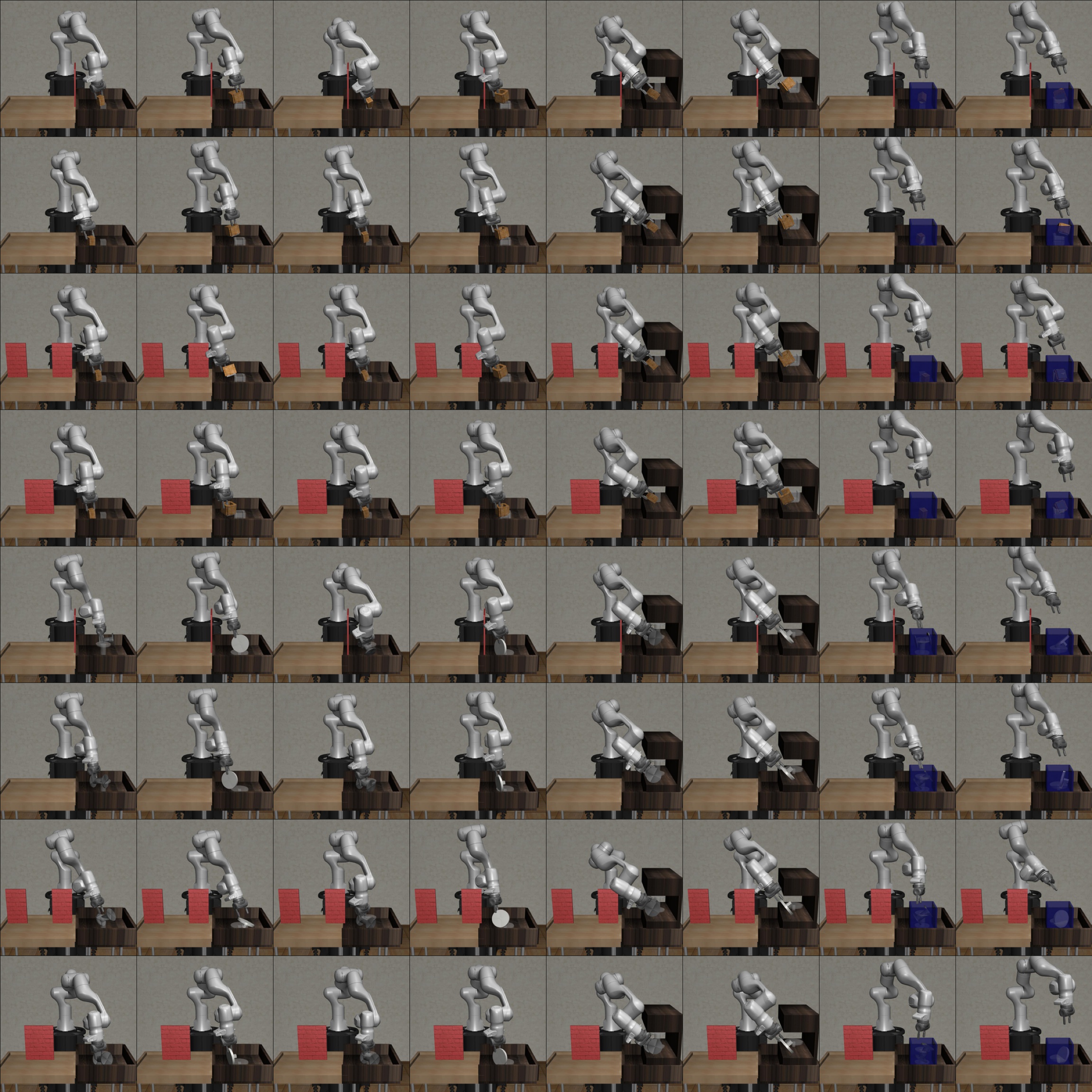}
    \caption{Visualization of the $64$ \Panda{} tasks.}
    \label{fig:panda_viz}
\end{figure}

\begin{figure}[p]
        \includegraphics[width=0.98\linewidth, trim={0.3cm 0.0cm 1.5cm 0.8cm}]{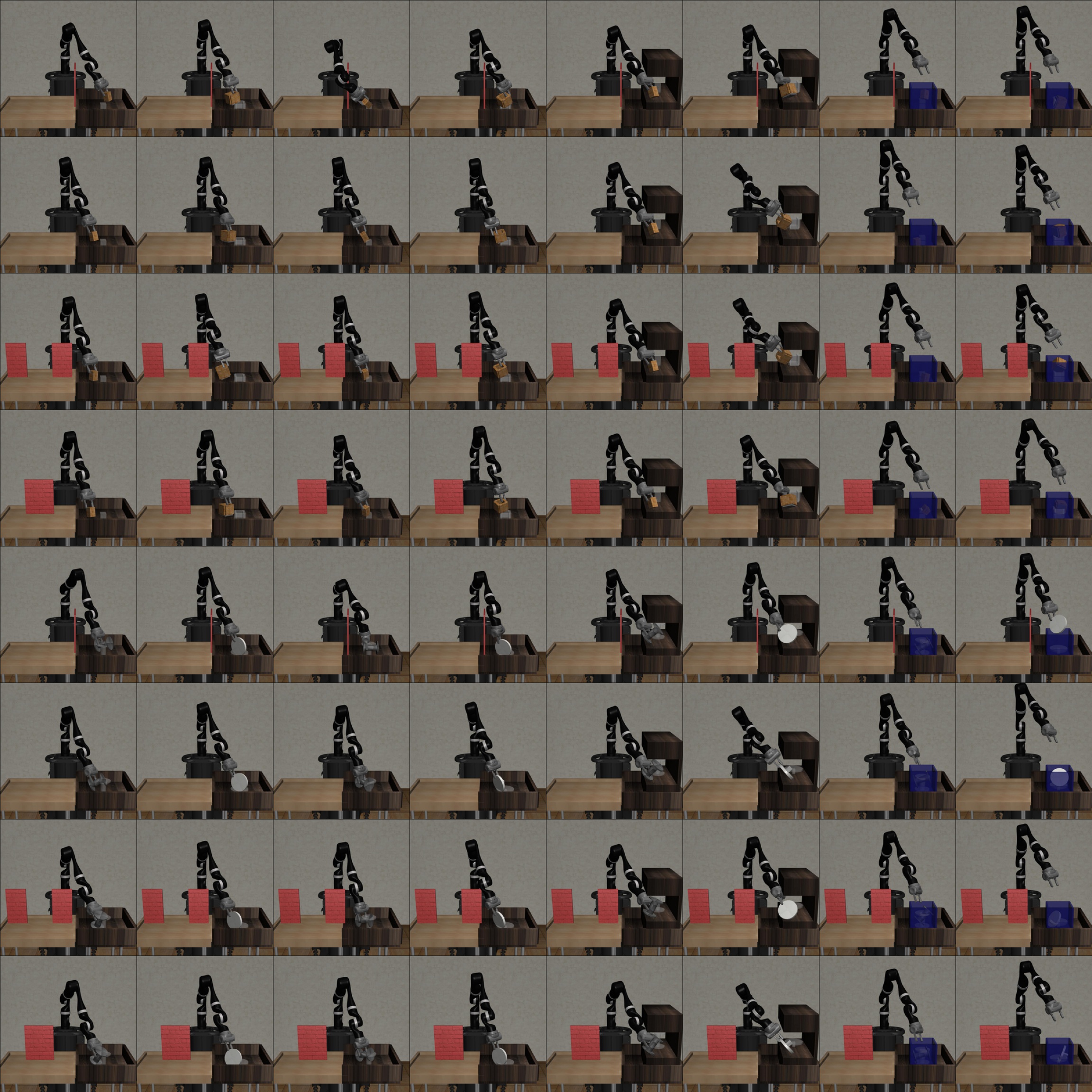}
    \caption{Visualization of the $64$ \Jaco{} tasks.}
    \label{fig:jaco_viz}
\end{figure}

\begin{figure}[p]
        \includegraphics[width=0.98\linewidth, trim={0.3cm 0.0cm 1.5cm 0.8cm}]{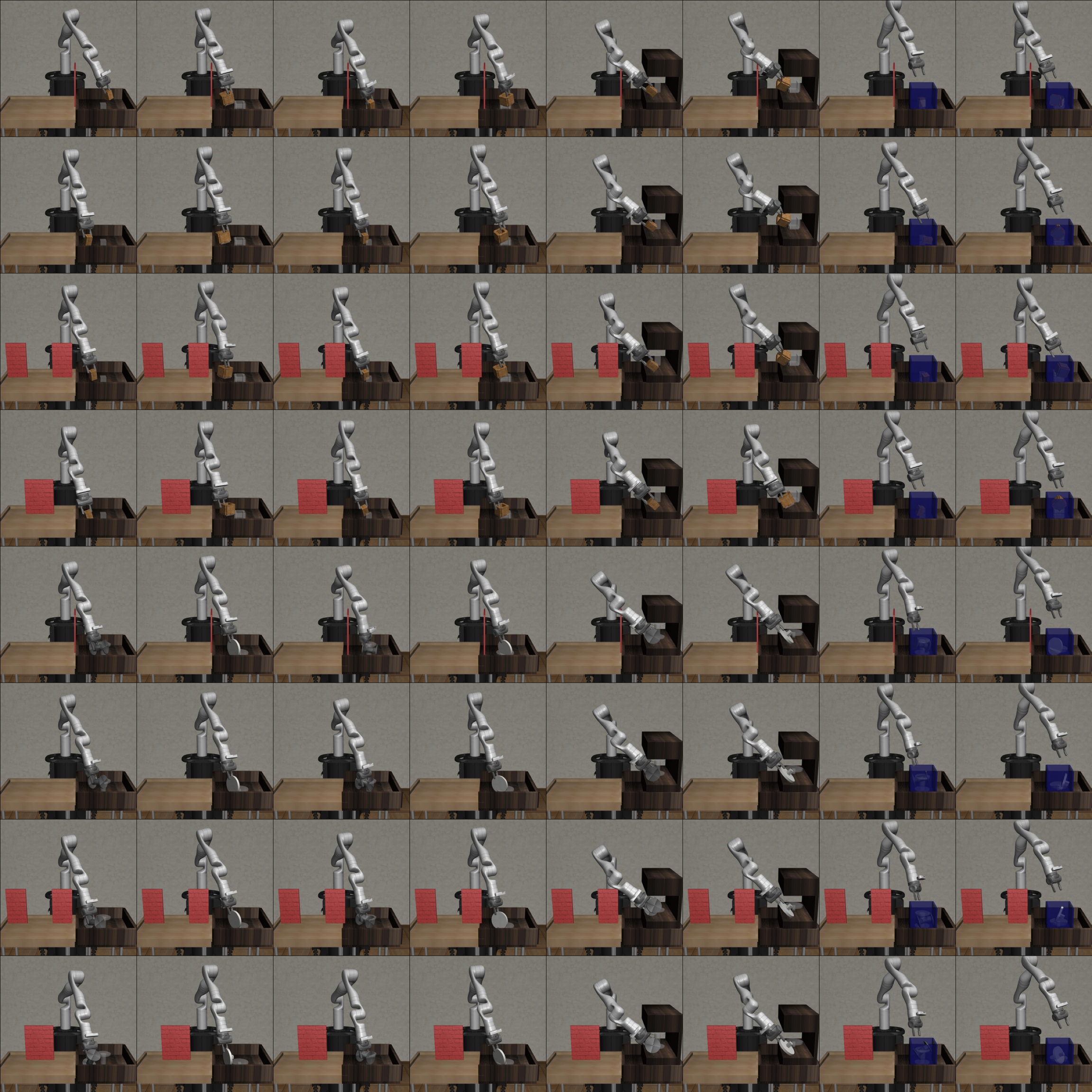}
    \caption{Visualization of the $64$ \Kinova{} tasks.}
    \label{fig:kinova_viz}
\end{figure}

\benchmark{} consists of a total of $256$ possible combinations of elements, each representing a separate task. Figures~\ref{fig:iiwa_viz}--~\ref{fig:kinova_viz} show each of the different robot arms in action solving the enormous diversity of tasks in \benchmark{}. 

\clearpage
\section{Reward functions}
\label{app:RewardFunctions}

Section~\ref{sec:RewardFunctions} in the main paper describes at a high level the structure of the reward function used for each task. Here, we include the precise mathematical formulas used to calculate them. In particular, the reward function computation depended only on the task objective and not any of the other axes.

\begingroup
\allowdisplaybreaks
\subsection{\pickPlace{} tasks}
\begin{flalign*}
    \Reward_{\mathrm{reach}} =& 0.2\, ( 1 - \tanh(10\cdot \mathrm{target\_dist}))&&&\\
    \Reward_{\mathrm{grasp}} =& \begin{cases}
        0.3 \quad &\text{if grasping}\\
        0 \quad &\text{otherwise}
    \end{cases}
    &&\\
    \Reward_{\mathrm{lift}} =& \begin{cases}
        0.3 + 0.2\, (1 - \tanh(5\cdot \mathrm{z\_dist\_target\_height})) \quad &\text{if }\Reward_{\mathrm{grasp}} > 0\\
        0 \quad &\text{otherwise}
    \end{cases}&&\\
    \Reward_{\mathrm{approach}} =& \begin{cases}
        \Reward_{\mathrm{lift}} + 0.2\, (1 - \tanh(2\cdot \mathrm{goal\_xy\_dist})) \quad &\text{if } \Reward_{\mathrm{lift}} > 0.45 \text{ and object is not above bin}\\
        0.5 + 0.2\, (1 - \tanh(2\cdot \mathrm{goal\_xy\_dist})) \quad &\text{if } \Reward_{\mathrm{lift}} > 0.45 \text{ and object is above bin}\\
        0 \quad &\text{otherwise}
    \end{cases}&&\\
    \Reward_{\mathrm{lower}} =& \begin{cases}
        0.7 + 0.2\, (1-\tanh(5\cdot\mathrm{z\_dist\_bin})) \quad &\text{if object is above bin and } \Reward_{\mathrm{grasp}} > 0\\
        0 \quad &\text{otherwise}
    \end{cases}&&\\ 
    \Reward_{\mathrm{success}} =& \begin{cases}
        1 \quad &\text{if object is in bin and } \Reward_{\mathrm{reach}} > 0.07\\
        0 \quad &\text{otherwise}
    \end{cases}&&\\
    \Reward =& \max_{\mathrm{stage}}\Reward_{\mathrm{stage}}&&
\end{flalign*}
    
\subsection{\push{} tasks}

\begin{flalign*}
    \Reward_{\mathrm{reach}} =& 0.2\, ( 1 - \tanh(10\cdot \mathrm{target\_dist}))&&\\
    \Reward_{\mathrm{grasp}} =& \begin{cases}
        0.3 \quad &\text{if grasping}\\
        0 \quad &\text{otherwise}
    \end{cases}&&
    \\
    \Reward_{\mathrm{approach}} =& \begin{cases}
        0.3 + 0.4\, (1 - \tanh(5\cdot \mathrm{goal\_xy\_dist})) \quad &\text{if }\Reward_{\mathrm{grasp}} > 0\\
        0 \quad &\text{otherwise}
    \end{cases}&&\\
    \Reward_{\mathrm{success}} =& \begin{cases}
        1 \quad &\text{if } \mathrm{goal\_xy\_dist} \leq 0.03\\
        0 \quad &\text{otherwise}
    \end{cases}&&\\
    \Reward =& \max_{\mathrm{stage}}\Reward_{\mathrm{stage}}&& 
\end{flalign*}

\subsection{\trashcan{} tasks}

\begin{flalign*}
    \Reward_{\mathrm{reach}} =& 0.2\, ( 1 - \tanh(10\cdot \mathrm{target\_dist}))&&\\
    \Reward_{\mathrm{grasp}} =& \begin{cases}
        0.3 \quad &\text{if grasping and object is not in trash can}\\
        0 \quad &\text{otherwise}
    \end{cases}&&
    \\
    \Reward_{\mathrm{lift}} =& \begin{cases}
        0.3 + 0.2\, (1 - \tanh(5\cdot \mathrm{z\_dist\_target\_height})) \quad &\text{if }\Reward_{\mathrm{grasp}} > 0 \text{ and object is not in trashcan}\\
        0 \quad &\text{otherwise}
    \end{cases}&&\\
    \Reward_{\mathrm{approach}} =& \begin{cases}
        \Reward_{\mathrm{lift}} + 0.2\, (1 - \tanh(2\cdot \mathrm{goal\_xy\_dist})) \quad &\text{if } \Reward_{\mathrm{lift}} > 0.45 \text{ and object is not in or above trash can}\\
        0.5 + 0.2\, (1 - \tanh(2\cdot \mathrm{goal\_xy\_dist})) \quad &\text{if } \Reward_{\mathrm{lift}} > 0.45 \text{ and object is above trash can}\\
        0 \quad &\text{otherwise}
    \end{cases}&&\\
    \Reward_{\mathrm{drop}} =& \begin{cases}
        0.95 \quad &\text{if object is above trashcan and } \Reward_{\mathrm{grasp}} = 0\\
        0 \quad &\text{otherwise}
    \end{cases}&&\\ 
    \Reward_{\mathrm{success}} =& \begin{cases}
        1 \quad &\text{if object is in trash can and gripper is not in trash can}\\
        0 \quad &\text{otherwise}
    \end{cases}&&\\
    \Reward =& \max_{\mathrm{stage}}\Reward_{\mathrm{stage}}&&
\end{flalign*}

\subsection{\shelf{} tasks}

\begin{flalign*}
    \Reward_{\mathrm{reach}} =& 0.2\, ( 1 - \tanh(10\cdot \mathrm{target\_dist}))&&\\
    \Reward_{\mathrm{grasp}} =& \begin{cases}
        0.3 \quad &\text{if grasping}\\
        0 \quad &\text{otherwise}
    \end{cases}&&
    \\
    \Reward_{\mathrm{lift}} =& \begin{cases}
        0.3 + 0.2\, (1 - \tanh(5\cdot \mathrm{z\_dist\_target\_height})) \quad &\text{if }\Reward_{\mathrm{grasp}} > 0\\
        0 \quad &\text{otherwise}
    \end{cases}&&\\
    \Reward_{\mathrm{align}} =& \begin{cases}
        0.5 + 0.3\, (1 - \tanh(\mathrm{y\_axis\_orientation})) \quad &\text{if object is in front of shelf}\\
        0 \quad &\text{otherwise}
    \end{cases}&&\\
    \Reward_{\mathrm{approach}} =& \begin{cases}
        0.8 + 0.1\, (1-\tanh(5\cdot\mathrm{y\_dist\_shelf})) \quad &\text{if object is in front of shelf and } \Reward_{\mathrm{align}} > 0.6\\
        0 \quad &\text{otherwise}
    \end{cases}&&\\ 
    \Reward_{\mathrm{success}} =& \begin{cases}
        1 \quad &\text{if object is in shelf}\\
        0 \quad &\text{otherwise}
    \end{cases}&&\\
    \Reward =& \max_{\mathrm{stage}}\Reward_{\mathrm{stage}}&&
\end{flalign*}
\endgroup

\section{Experimental details}
\label{app:PPODetails}

This section provides details of the experimental setting used to obtain all results in Section~\ref{sec:experiments} in the main paper.

\subsection{PPO details and hyper-parameters}

We used a modified version of \texttt{Spinning\! Up}'s PPO implementation for all experiments. These changes were made to encourage improved exploration because, despite the introduction of highly crafted rewards, initial experiments with the original implementation were suffering from premature convergence. 

In particular, we used a multi-layer perceptron (MLP) to represent the mean of a Gaussian policy. Against popular wisdom, which encourages using linear activations in the final layer, we found that adding a $\tanh$ activation led to substantially improved exploration. The rationale is that robot actions are typically capped (artificially in simulators, and by physical limits in real robots). Therefore, if the MLP outputs high-magnitude means for the Gaussian distribution, the sampled actions are all likely to reach the range limits, regardless of the variance of the Gaussian. In consequence, the agent could ``cheat'' existing techniques to avoid premature convergence (e.g., entropy regularization) by learning a high variance but being deterministic in practice by saturating the actions. The $\tanh$ activation ensures that the actions are never too large in magnitude, which permits the sampling to induce stochasticity (and, consequently, exploration).

The second modification was to use a constant variance for the Gaussian policy, instead of propagating gradients through it. The reason this was necessary is that, with a learnable variance, the agent was finding pathological regions of the optimization landscape that (once more) cheated existing entropy regularization approaches. Concretely, the agent was inflating the variance along dimensions where actions were inconsequential (e.g., joints that rotate in directions orthogonal to the motion of the gripper), and reducing the variance to a minimum along critical dimensions. The resulting policy was therefore deterministic along all interesting dimensions, and so the exploration the agent was engaging in was ineffective. Setting a fixed variance of $\sigma^2=1$ ($\log(\sigma) = 0$) for the seven joint actions and $\sigma^2= 1/e$ ($\log(\sigma)=-0.5$) for the gripper action ensured that the robot consistently explored throughout the learning process and was critical toward enabling the agent to learn the \benchmark{} tasks.

The hyper-parameters used for training, reported in Table~\ref{tab:hyperparams}, were obtained via grid search on a set of tasks using single-task training and maintained for the multi-task and compositional settings.

\begin{table}[t!]
    \centering
    \caption{PPO hyper-parameters used to train all agents, obtained via grid search with the single-task agent.}
    \vspace{-0.5em}
    \begin{tabular}{l|c|c|c}
         & Single-Task Learner & Multi-Task Learner & Compositional Learner \\
         \hline
        $\gamma$ & $0.99$ & $0.99$ & $0.99$ \\
        \# layers & $2$ & $2$ & --- \\
        \# hidden units & $64$ & $256$ & --- \\
        \# steps per task per update & $16,\!000$ & $16,\!000$ & $16,\!000$ \\
        \# total step per task & $10,\!000,\!000$ & $10,\!000,\!000$ & $10,\!000,\!000$ \\
        PPO clip value & $0.2$ & $0.2$ & $0.2$ \\
        $\pi$ learning rate & $1\cdot10^{-4}$ & $1\cdot10^{-4}$ & $1\cdot10^{-4}$ \\
        $V$ learning rate & $1\cdot10^{-4}$ & $1\cdot10^{-4}$ & $1\cdot10^{-4}$ \\
        \# $\pi$ update iterations & $128$ & $128$ & $128$ \\
        \# $V$ update iterations & $128$ & $128$ & $128$ \\
        Target KL & $0.02$ & $0.02$ & $0.02$ \\
    \end{tabular}
    \label{tab:hyperparams}
\end{table}

\subsection{Compositional network architecture}

\begin{figure}[t!]
\centering
    \hspace{0.16\textwidth}
    \centering
        \includegraphics[width=0.8\linewidth, trim={2cm 3.5cm 3cm 4.5cm}, clip]{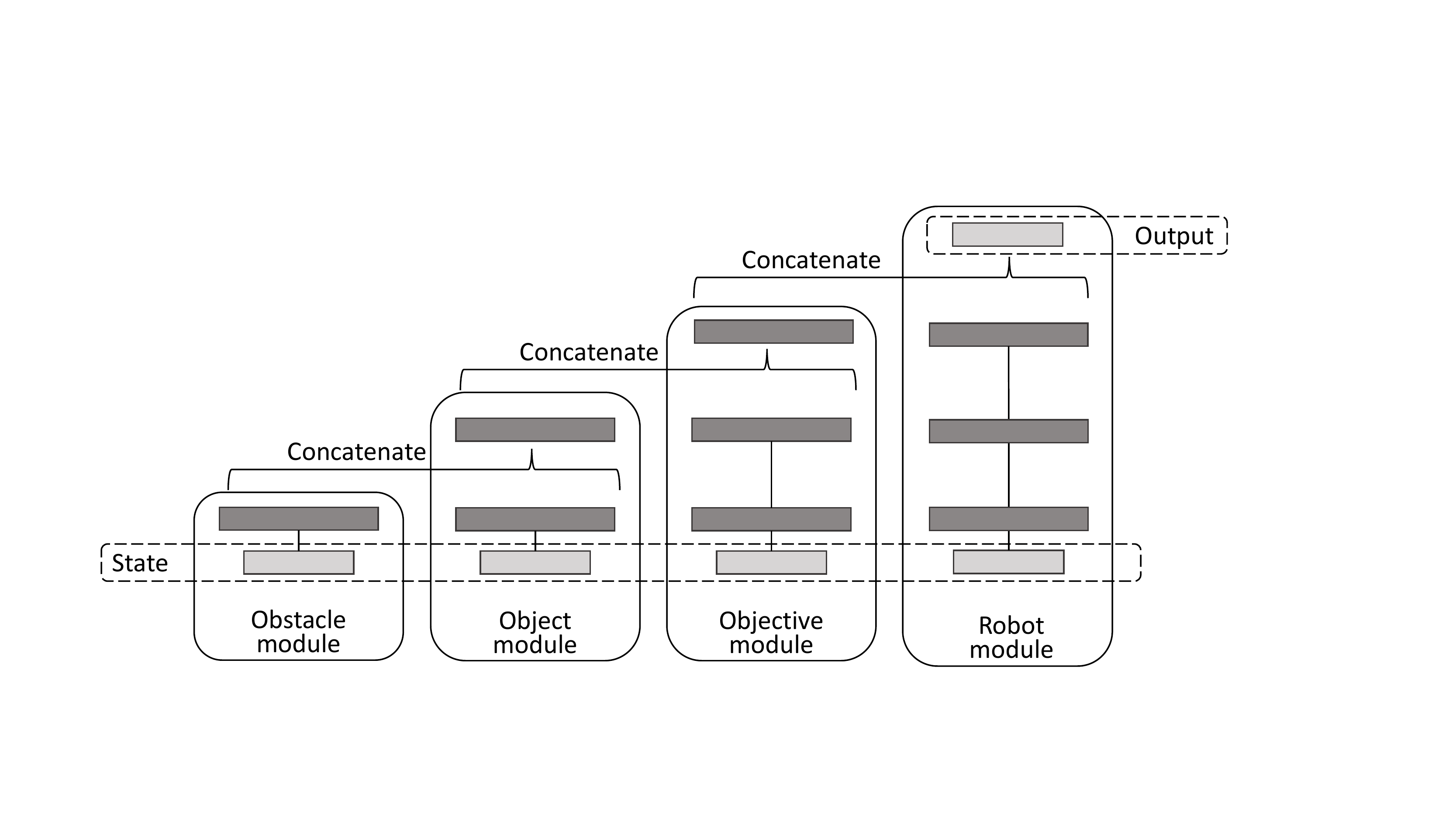}
    \vspace{-0.5em}
    \caption{Modular architecture used for learning compositional policies.}
    \label{fig:architecture}
\end{figure}

The network architecture for the compositional agent (Figure~\ref{fig:architecture}) follows a graph structure similar to that proposed by~\citet{mendez2022modular}. The network consists of a total of $16$ modules, each of which is represented by an MLP and corresponds to one of the compositional elements in \benchmark{}. More specifically, there are four obstacle, object, objective, and robot modules, respectively. The modules are assigned to levels in a graph hierarchy, such that the previous level's MLP output is concatenated with the output of the second-to-last layer of the current level's MLP. The concatenated tensor is then used as input to the final layer of the MLP of the current level. Concretely, the obstacle observation is processed first. Every obstacle module consists of a single-hidden-layer MLP with $32$ hidden units---because this is the first level, there is no additional input other than the obstacle observation. The second input that is processed is the object input. Object modules consist of two hidden layers, each of size $32$. The output of the obstacle network is concatenated with the output of the first layer of the object module and used as input to the second layer of the object module. The object module feeds into the second layer of the objective module, which consists of three layers of size $64$. Finally, the objective module's output is routed into the third layer of the robot module, which has a total of four layers: the first three layers are of size $64$, and the final layer is the output layer. The same network architecture was used to model both the value function and the policy. This architecture was selected to approximately match the total number of parameters in the multi-task network architecture.

\subsection{Computational resources}

We ran experiments using multiple AMD EPYC\textsuperscript{\texttrademark} central processing units (CPUs) with $128$-thread support. To parallelize data collection, each experiment was run across several processes using MPI with one CPU core per two MPI processes. Every process corresponded to an additional environment collecting samples. The single-task experiments used $16$ parallel processes running environments of the same task. Each single-task experiment had an approximate wall-clock training time of $12$ hours. For multi-task and compositional training, a single process per task was used. For an experiment of $56$ tasks, the agent was trained via $56$ MPI processes for approximately $4$ days. For an experiment of $224$ tasks, the training on $224$ processes took approximately $12$ days. There was no significant run-time difference between multi-task and compositional training. In general, we allocated roughly $1$ GB of RAM per MPI process.

\begin{figure}[p]
\captionsetup[subfigure]{font=scriptsize, aboveskip=2pt}
\centering
    \begin{subfigure}[b]{0.7\textwidth}
            \centering
                \includegraphics[width=\linewidth]{figures/legend.pdf}
        \end{subfigure}\\
    \begin{subfigure}[b]{0.3\textwidth}
    \centering
        \includegraphics[height=3cm, trim={0.2cm 0.0cm 1.5cm 0.8cm}, clip]{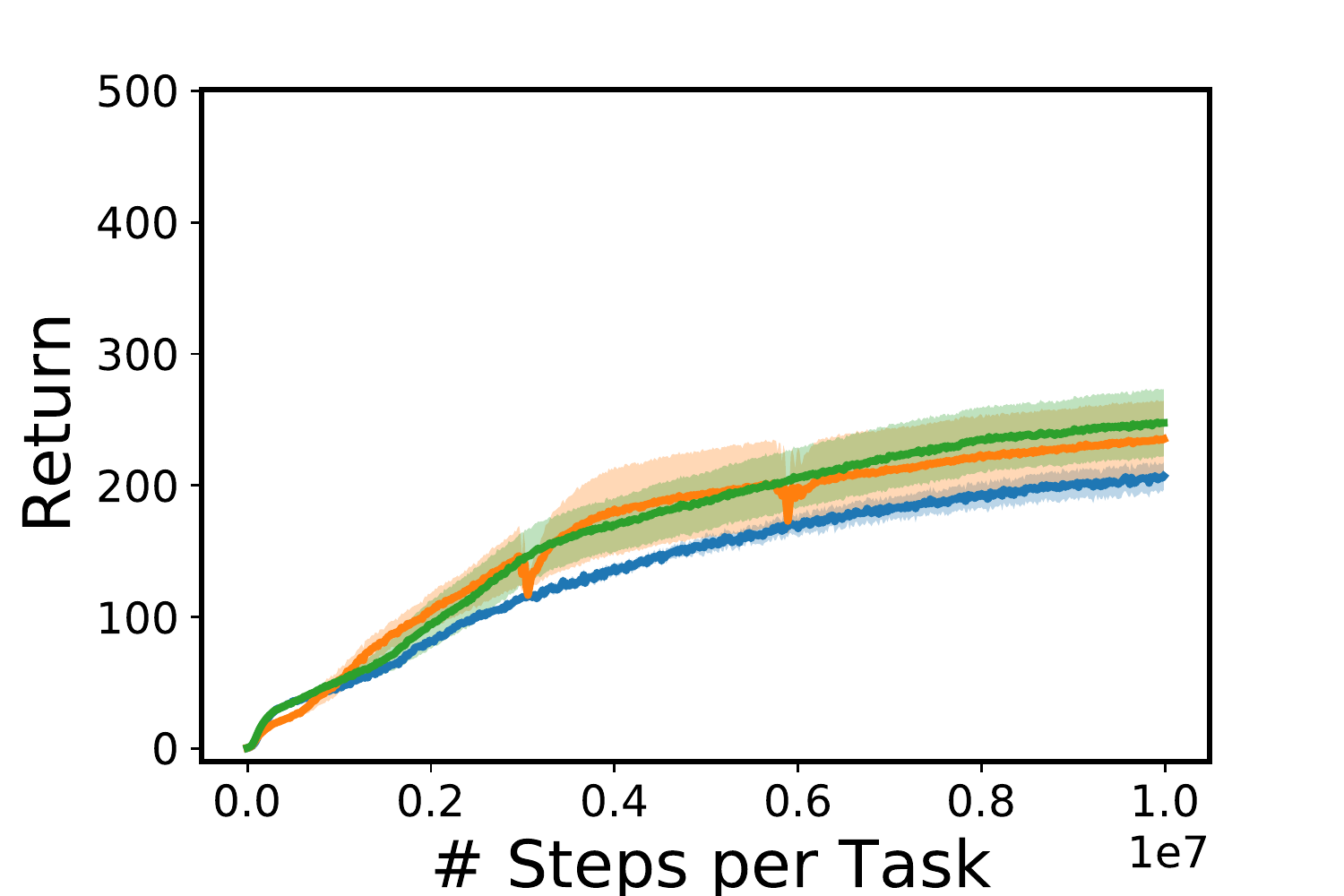}\\
        \vspace{-8.5em}
        {\hspace{20pt}$32$ training tasks}
    \end{subfigure}%
    \begin{subfigure}[b]{0.3\textwidth}
    \centering
        \includegraphics[height=3cm, trim={2.2cm 0.0cm 1.5cm 0.8cm}, clip]{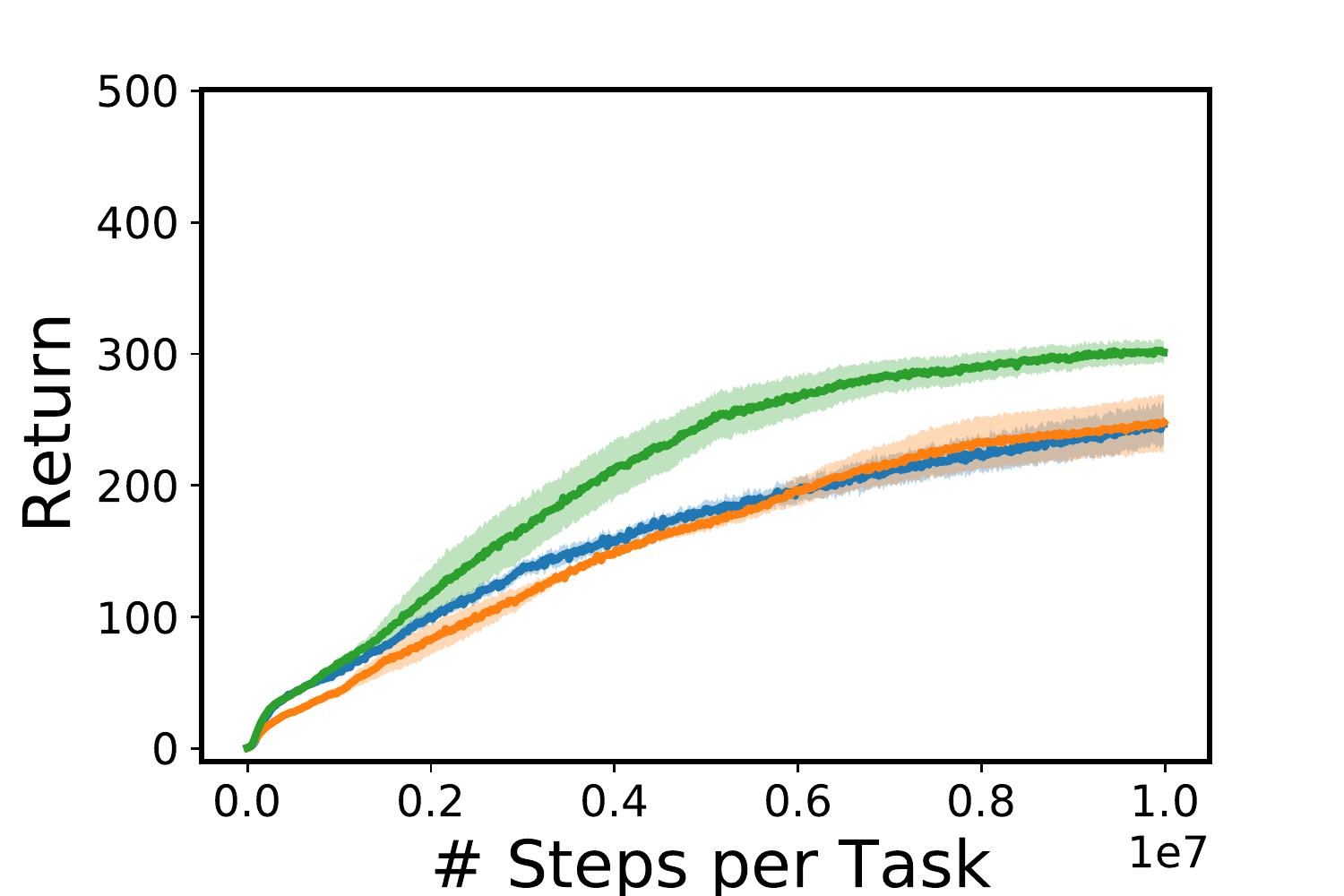}\\
        \vspace{-8.5em}
        {$32$ training tasks}
    \end{subfigure}%
    \begin{subfigure}[b]{0.3\textwidth}
    \centering
        \includegraphics[height=3cm, trim={2.2cm 0.0cm 1.5cm 0.8cm}, clip]{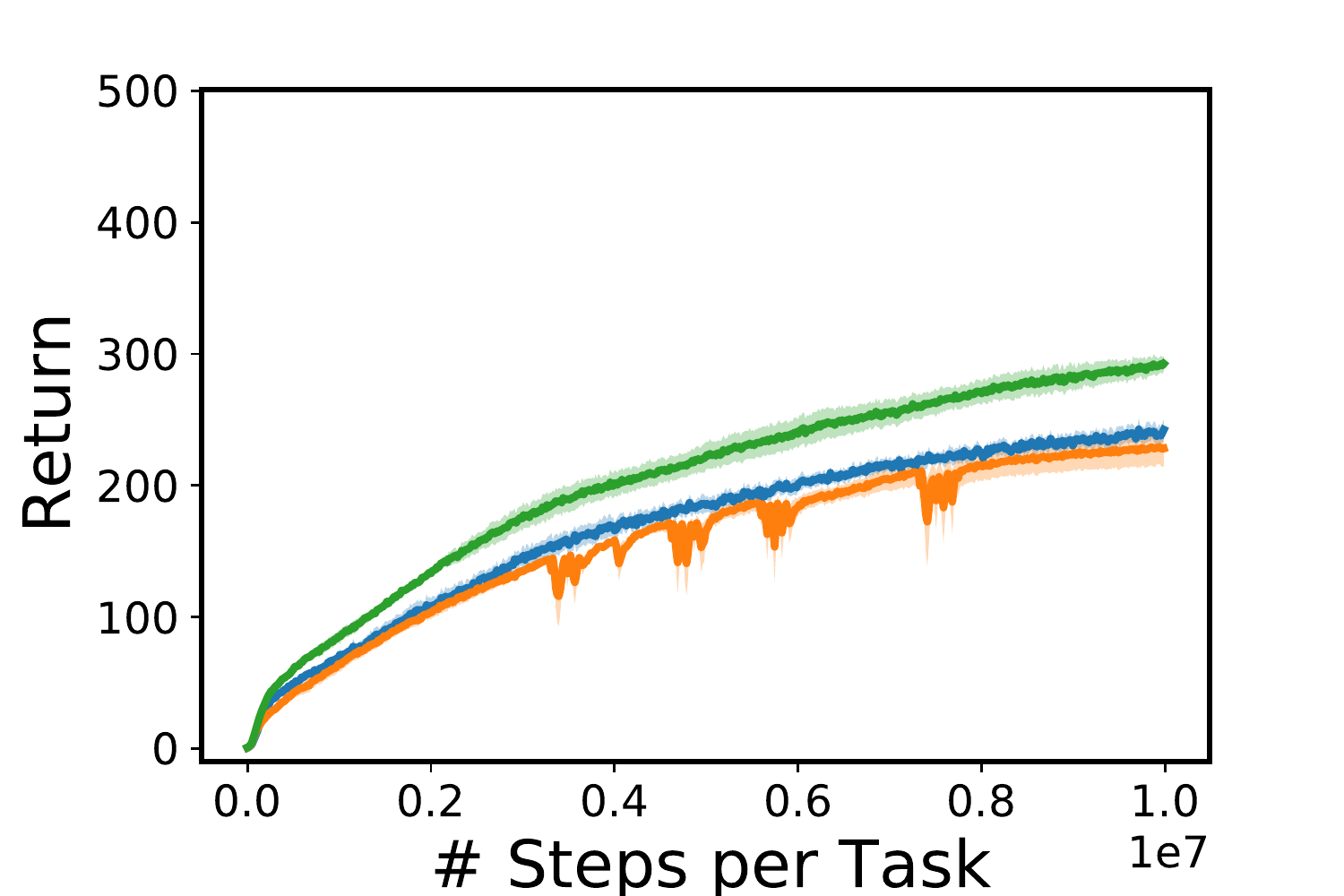}\\
        \vspace{-8.5em}
        {$32$ training tasks}
    \end{subfigure}%
    \vspace{7.1em}
    \\
    \begin{subfigure}[b]{0.3\textwidth}
    \centering
        \includegraphics[height=3cm, trim={0.2cm 0.0cm 1.5cm 0.8cm}, clip]{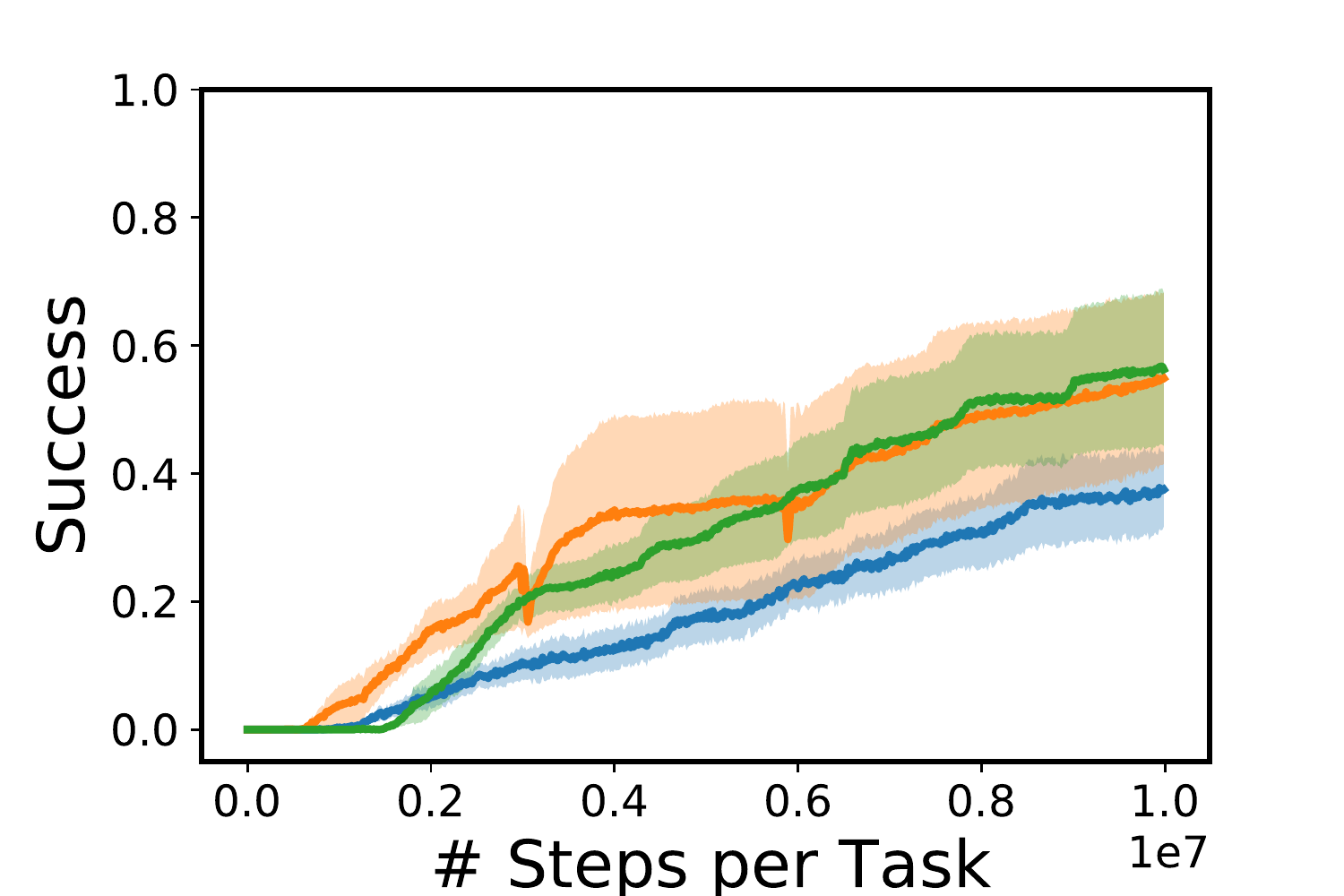}
        \caption{\benchmark{}$\cap$\hollowBox{}}
        \label{fig:smallscaleHollowbox}
    \end{subfigure}%
    \begin{subfigure}[b]{0.3\textwidth}
    \centering
        \includegraphics[height=3cm, trim={2.2cm 0.0cm 1.5cm 0.8cm}, clip]{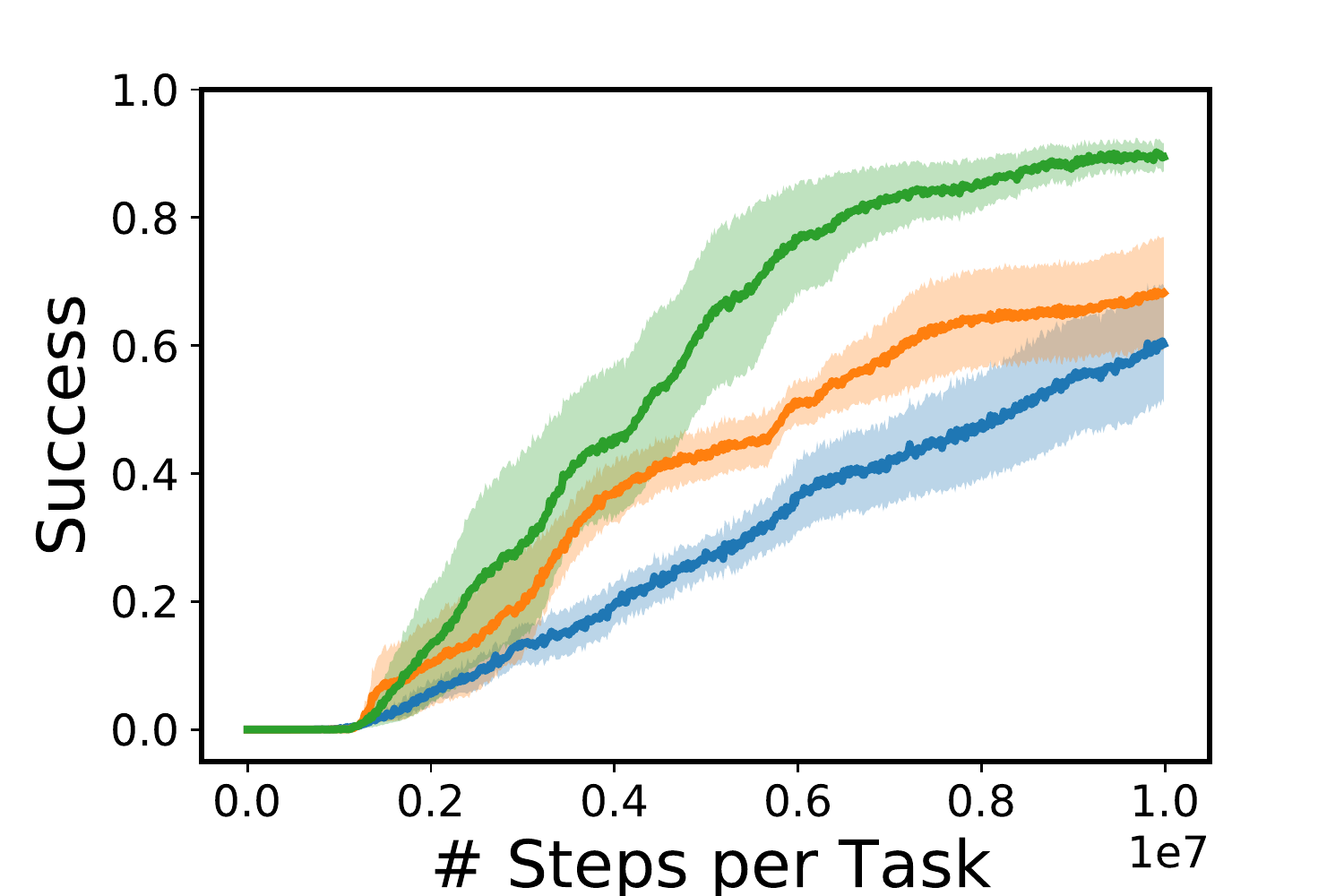}
        \caption{\benchmark{}$\cap$\pickPlace{}}
        \label{fig:smallscalePickPlace}
    \end{subfigure}%
    \begin{subfigure}[b]{0.3\textwidth}
    \centering
        \includegraphics[height=3cm, trim={2.2cm 0.0cm 1.5cm 0.8cm}, clip]{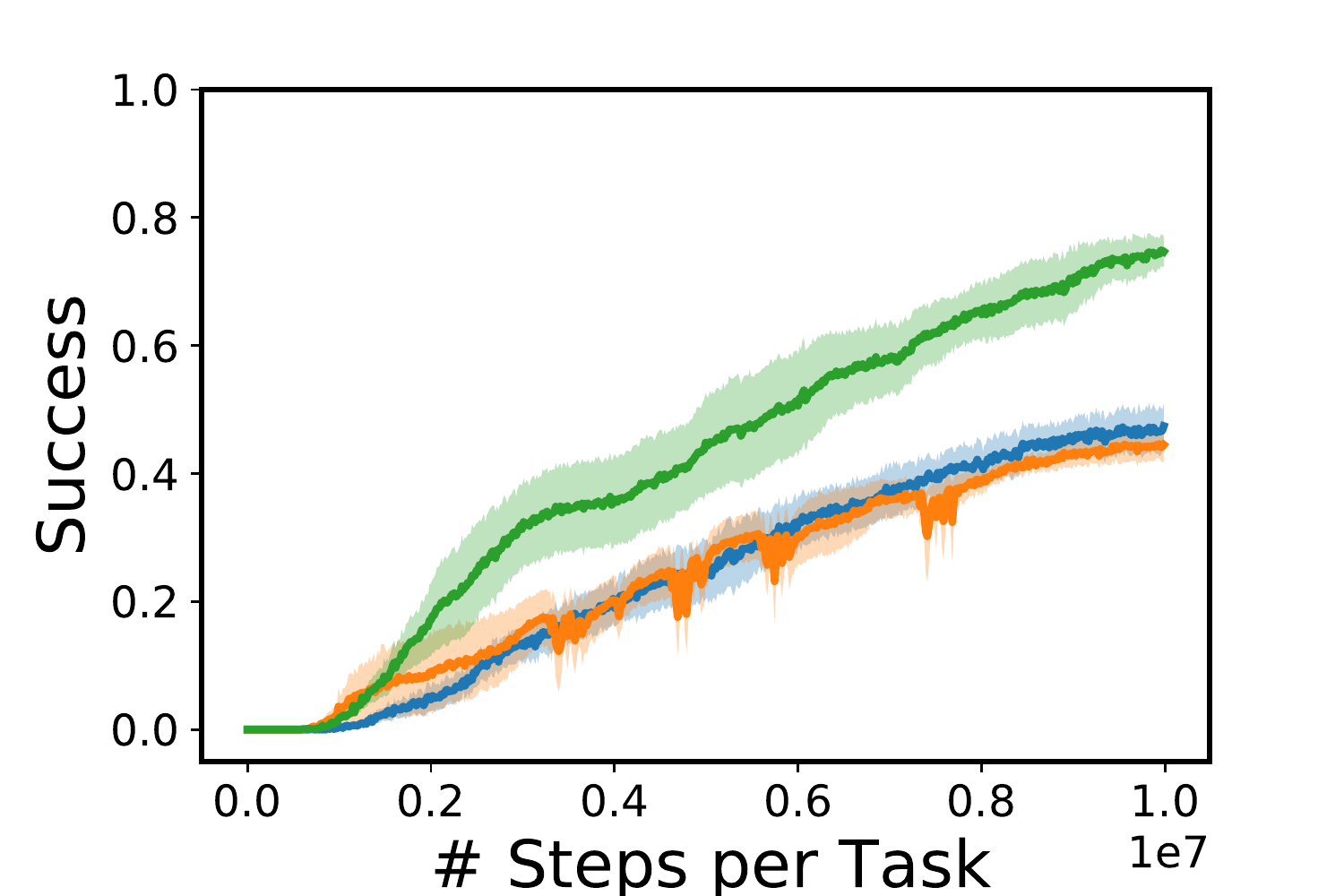}
        \caption{\benchmark{}$\cap$\noObstacle{}}
        \label{fig:smallscaleNoObstacle}
    \end{subfigure}%
    \vspace{-1em}
    \caption{Training performance on remaining suggested smaller-scale experiments. The compositional agent outperformed the competing baselines when all tasks involved \pickPlace{} or \noObstacle{}, and matched the multi-task agent when they included \hollowBox{}.  Shaded regions represent standard errors across three seeds.}
    \label{fig:additionalSmallscale}
\end{figure}

\begin{figure}[p]
\captionsetup[subfigure]{font=scriptsize, aboveskip=2pt}
\centering
    \begin{subfigure}[b]{0.7\textwidth}
            \centering
                \includegraphics[width=\linewidth]{figures/legend.pdf}
        \end{subfigure}\\
    \begin{subfigure}[b]{0.3\textwidth}
    \centering
        \includegraphics[height=3cm, trim={0.2cm 0.0cm 1.5cm 0.8cm}, clip]{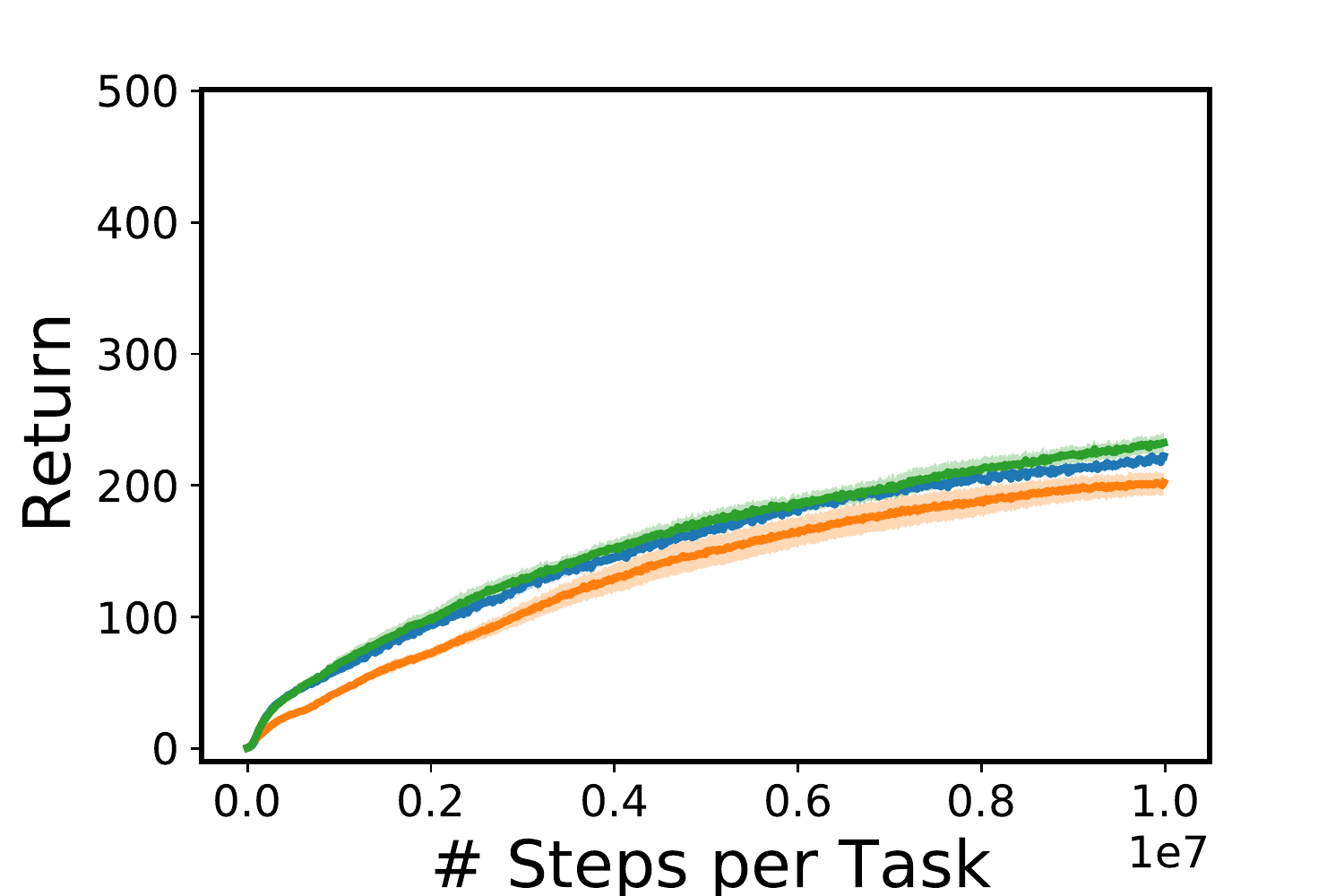}\\
        \vspace{-8.5em}
        {\hspace{20pt}$56$ training tasks}
    \end{subfigure}%
    \begin{subfigure}[b]{0.3\textwidth}
    \centering
        \includegraphics[height=3cm, trim={2.2cm 0.0cm 1.5cm 0.8cm}, clip]{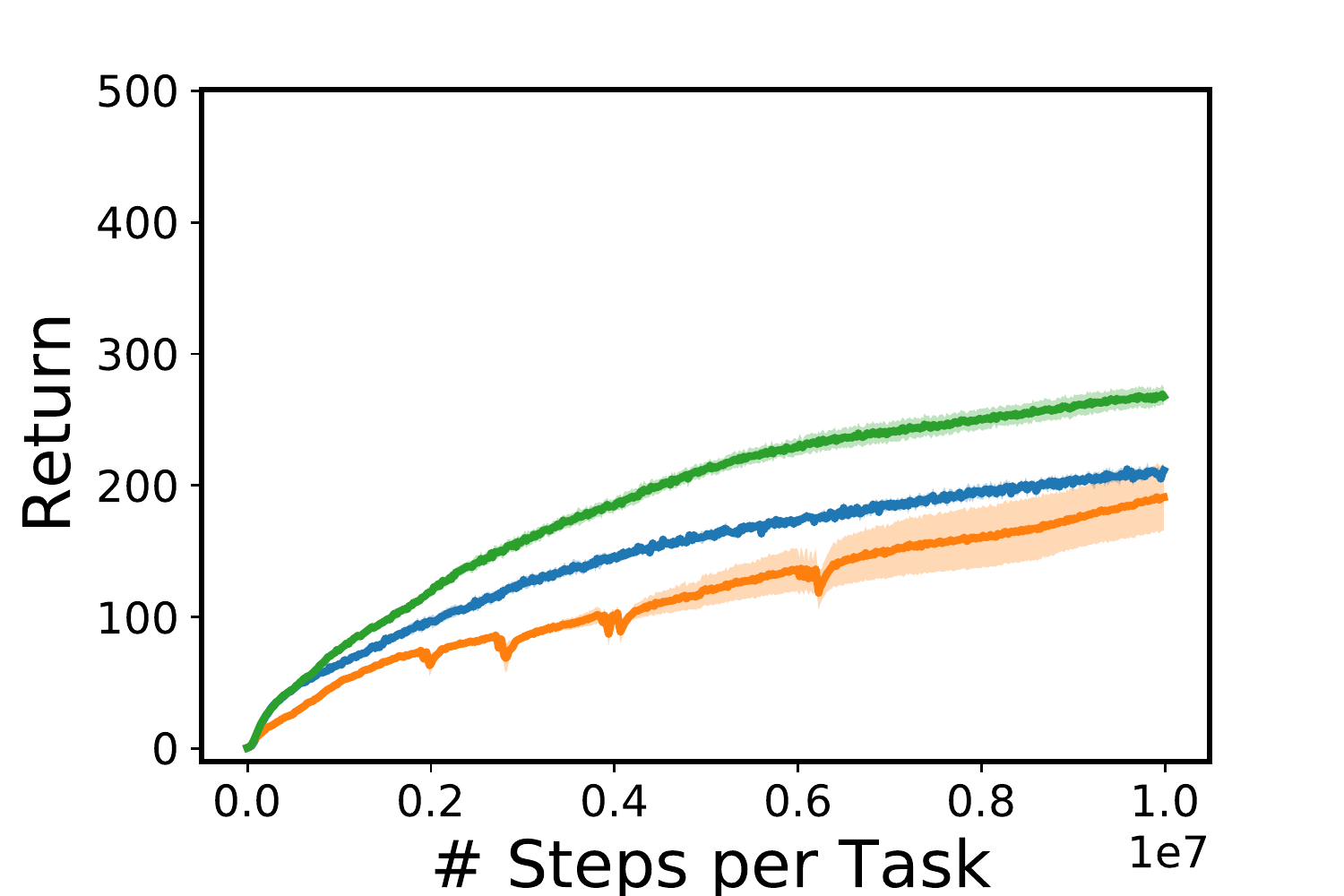}\\
        \vspace{-8.5em}
        {$56$ training tasks}
    \end{subfigure}%
    \begin{subfigure}[b]{0.3\textwidth}
    \centering
        \includegraphics[height=3cm, trim={2.2cm 0.0cm 1.5cm 0.8cm}, clip]{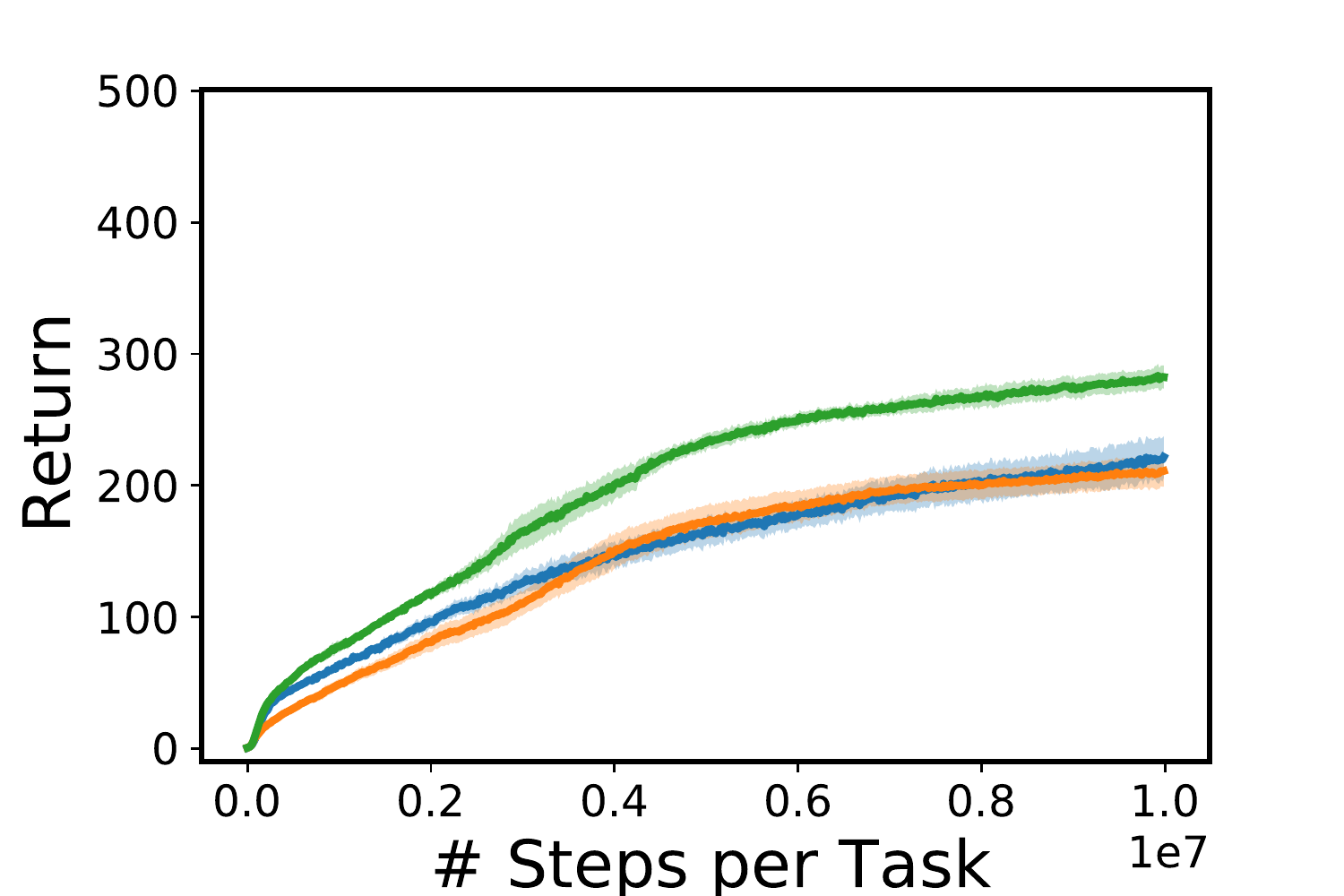}\\
        \vspace{-8.5em}
        {$56$ training tasks}
    \end{subfigure}%
    \vspace{7.1em}
    \\
    \begin{subfigure}[b]{0.3\textwidth}
    \centering
        \includegraphics[height=3cm, trim={0.2cm 0.0cm 1.5cm 0.8cm}, clip]{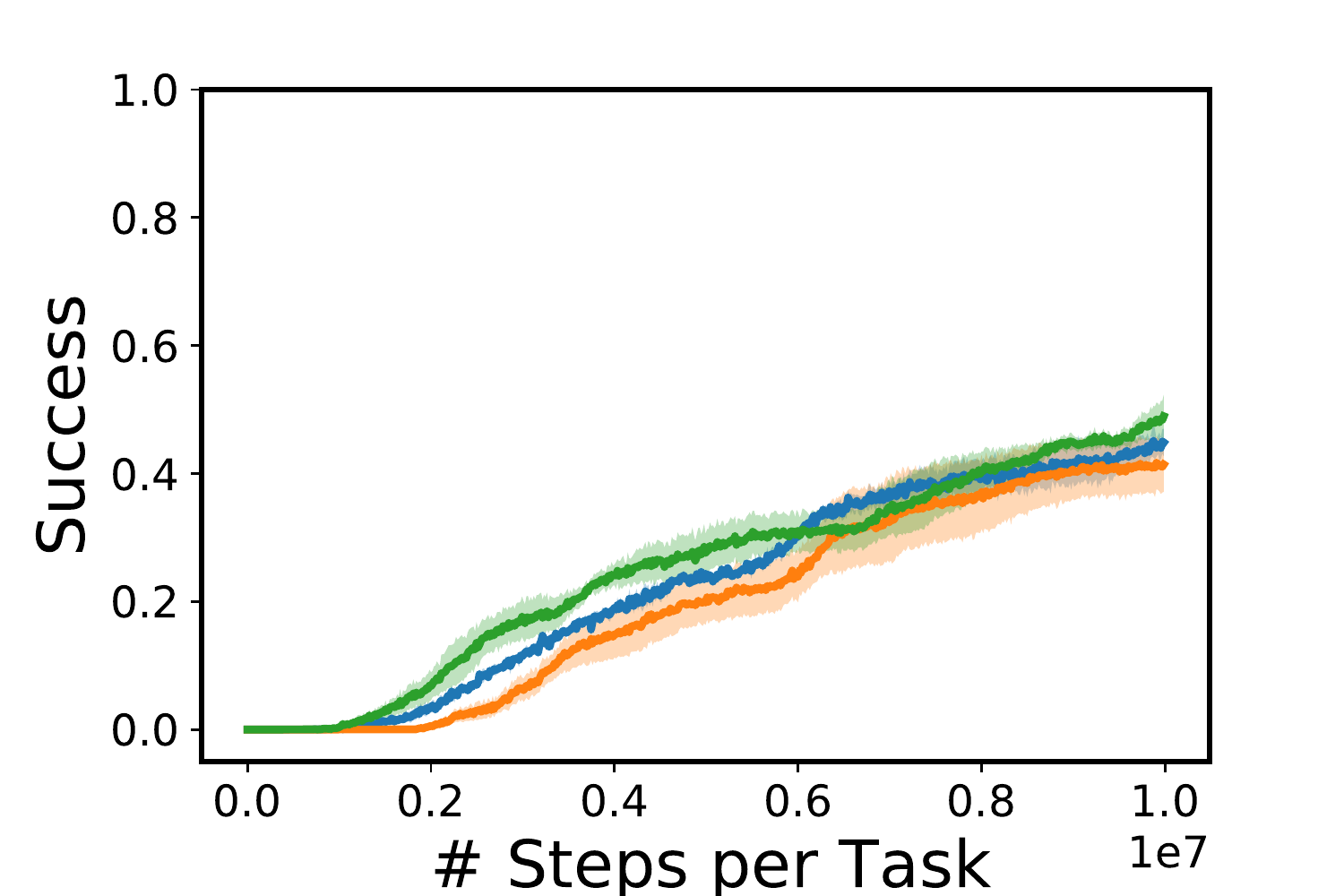}
        \caption{\benchmark{}\textbackslash{}\IIWA{}}
        \label{fig:holdoutIIWA}
    \end{subfigure}%
    \begin{subfigure}[b]{0.3\textwidth}
    \centering
        \includegraphics[height=3cm, trim={2.2cm 0.0cm 1.5cm 0.8cm}, clip]{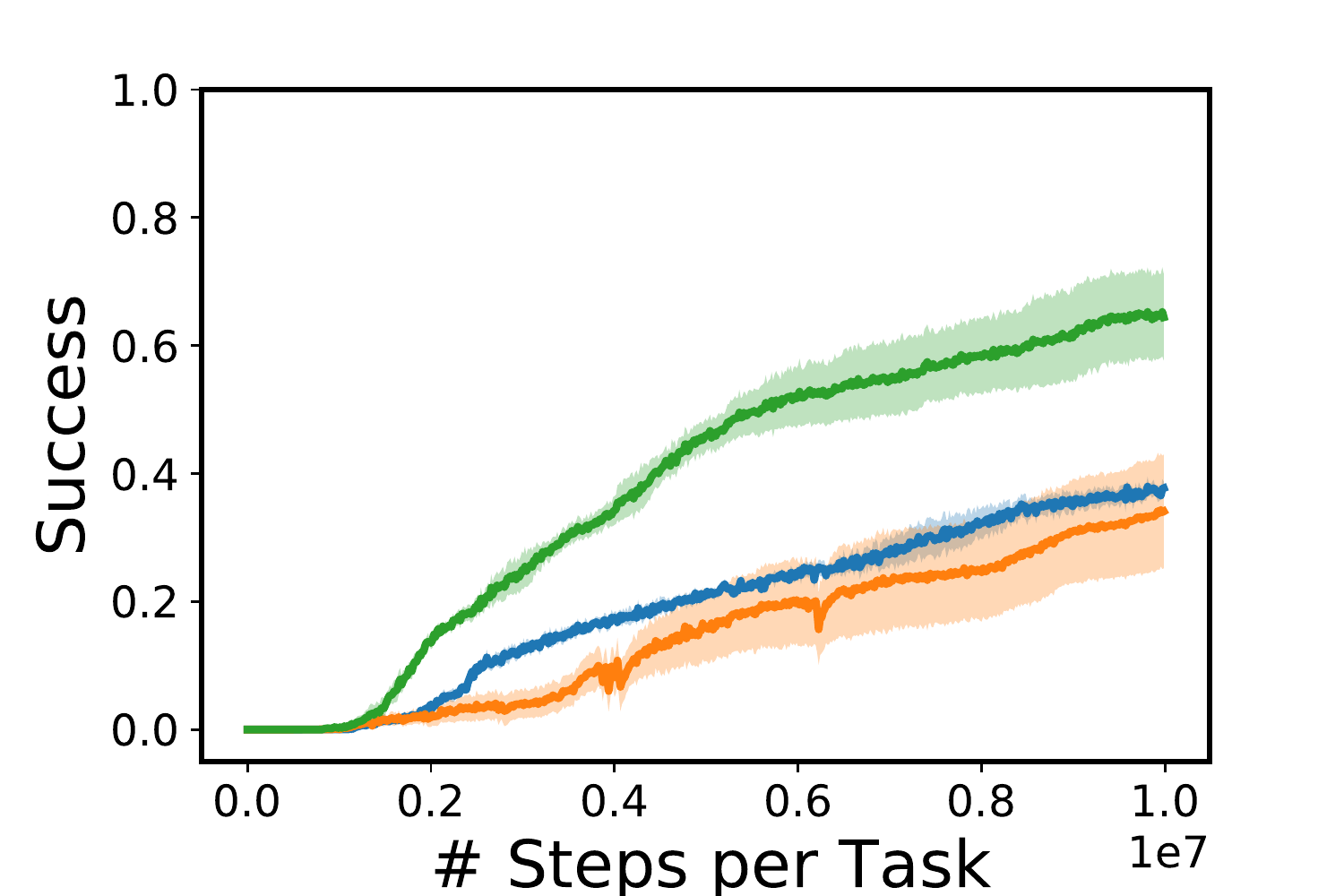}
        \caption{\benchmark{}\textbackslash{}\hollowBox{}}
        \label{fig:holdoutHollowbox}
    \end{subfigure}%
    \begin{subfigure}[b]{0.3\textwidth}
    \centering
        \includegraphics[height=3cm, trim={2.2cm 0.0cm 1.5cm 0.8cm}, clip]{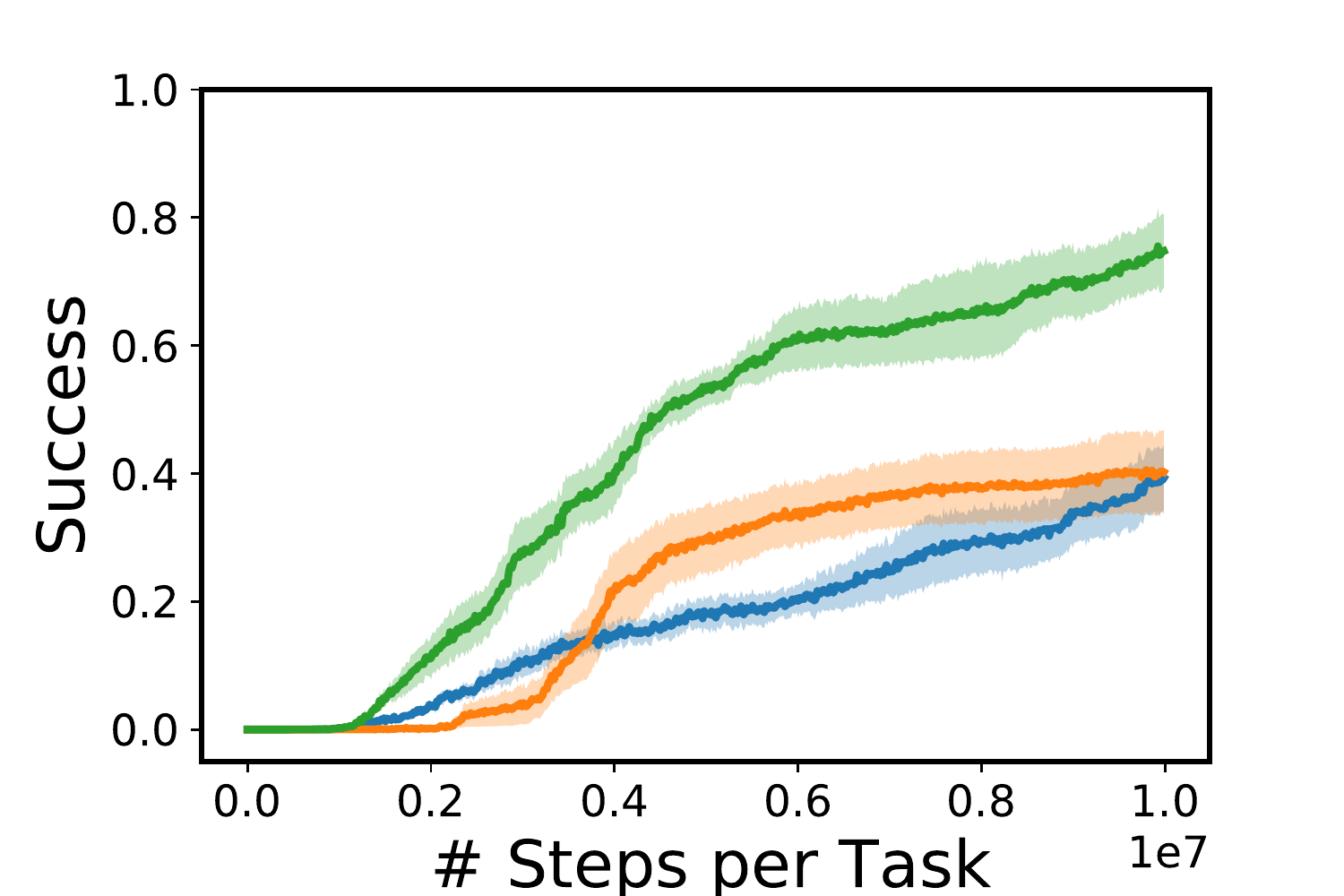}
        \caption{\benchmark{}\textbackslash{}\objectWall{}}
        \label{fig:holdoutObjectwall}
    \end{subfigure}%
    \vspace{-1em}
    \caption{Training performance on remaining suggested restricted sampling experiments. Performance of single-task and multi-task agents resembled that on the full benchmark, while the compositional agent performed better when trained on a single \hollowBox{} or \objectWall{} task.  Shaded regions represent standard errors across three seeds.}
    \label{fig:additionalHoldout}
\end{figure}

\begin{table}[p]
\addtolength{\tabcolsep}{-0.1em}
    \captionof{table}{\benchmark{} zero-shot generalization on remaining suggested smaller-scale settings. Only the compositional agent on \benchmark{}$\cap$\pickPlace{} generalized. Standard errors across three seeds reported after the $\pm$.}
    \label{tab:addZeroshotSmallscale}
    \centering
    \vspace{-1em}
        \begin{tabular}{l|ll|ll|ll}
         & \multicolumn{2}{c|}{\benchmark{}} & 
         \multicolumn{2}{c|}{\benchmark{}} & 
         \multicolumn{2}{c}{\benchmark{}}  \\
         & \multicolumn{2}{c|}{$\cap$\hollowBox{}} &
         \multicolumn{2}{c|}{$\cap$\pickPlace{}} &
         \multicolumn{2}{c}{$\cap$\noObstacle{}} \\
         \hline\hline
         & Multi-task & Comp. & Multi-task & Comp. & Multi-task & Comp.\\
         \hline
        Return & $99.71 \pm 31.32$ & $77.64 \pm 9.55$ & $96.05 \pm 15.55$ & $133.27 \pm 30.20$ & $129.70 \pm 22.43$ & $75.19 \pm 12.33$ \\
        Success & $0.13 \pm 0.06$ & $0.12 \pm 0.05$ & $0.22 \pm 0.08$ & $0.40 \pm 0.09$ & $0.18 \pm 0.06$ & $0.09 \pm 0.04$ \\
        \end{tabular}
\end{table}

\begin{table}[p]
\addtolength{\tabcolsep}{-0.1em}
    \captionof{table}{\benchmark{} zero-shot generalization on the additionally suggested restricted settings. The agents achieved nearly zero generalization. Standard errors across three seeds reported after the $\pm$.}
    \label{tab:addZeroshotHoldout}
    \centering
    \vspace{-1em}
        \begin{tabular}{l|ll|ll|ll}
         & \multicolumn{2}{c|}{\benchmark{}} & 
         \multicolumn{2}{c|}{\benchmark{}} & 
         \multicolumn{2}{c}{\benchmark{}} \\
         & \multicolumn{2}{c|}{\textbackslash{}\IIWA{}} & 
         \multicolumn{2}{c|}{\textbackslash{}\hollowBox{}}& 
         \multicolumn{2}{c}{\textbackslash{}\objectWall{}}\\
         \hline\hline
         & Multi-task & Comp. & Multi-task & Comp. & Multi-task & Comp.\\
         \hline
        Return & $30.62 \pm 7.52$ & $24.32 \pm 1.22$ & $74.53 \pm 42.78$ & $14.78 \pm 3.72 $ & $30.33 \pm 8.07$ & $9.88 \pm 2.73$ \\
        Success & $0.00 \pm 0.00$ & $0.00 \pm 0.01$ & $0.09 \pm 0.12$ & $0.01 \pm 0.01$ & $0.03 \pm 0.03$ & $0.01 \pm 0.01$ \\
        \end{tabular}
\end{table}

\clearpage

\section{Additional results on smaller-scale and restricted settings} \label{app:additionalSampling}

This section provides results on the remaining settings suggested in Section~\ref{sec:sampleTrainingTasks} in the main paper. Results on the training tasks in the smaller-scale setting, summarized in Figure~\ref{fig:additionalSmallscale}, demonstrate that multi-task agents can more easily transfer knowledge across tasks that use a common robot arm or that share a common objective. In contrast, the compositional agent outperformed the single-task agent in all settings, demonstrating its ability to solve even the most highly varied sets of tasks. Training performance on the restricted setting, shown in Figure~\ref{fig:additionalHoldout}, exhibits similar trends to the full \benchmark{}, as expected. As exceptions, when restricting the compositional agent to a single \hollowBox{} or \objectWall{} task it outperformed the other methods.

Table~\ref{tab:addZeroshotSmallscale} shows the zero-shot performance in smaller-scale settings, where both agents achieved little generalization, except the compositional agent on the \pickPlace{} setting. In particular, the multi-task agent failed to generalize in all settings with varied robot arms, supporting the claim of difficulty of generalization across robotic manipulators. 
Table~\ref{tab:addZeroshotHoldout} reports zero-shot results in the restricted sampling setting, where both agents completely failed to generalize.

\section{Maximum success rate per task}
\label{app:MaxSuccessRatePerTask}

\begin{figure}[b!]
\centering
    \begin{subfigure}[b]{1.\textwidth}
        \centering
            \includegraphics[width=0.4\linewidth, trim={0.0cm 0.0cm 0.0cm 0.0cm},clip]{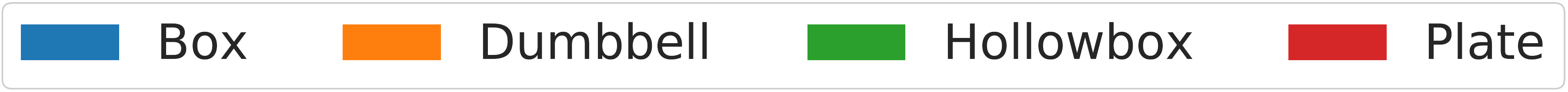}
    \end{subfigure}\\
    \begin{subfigure}[b]{1.\textwidth}
        \centering
            \includegraphics[width=0.75\linewidth, trim={0.3cm 0.0cm 1.5cm 1.4cm},clip]{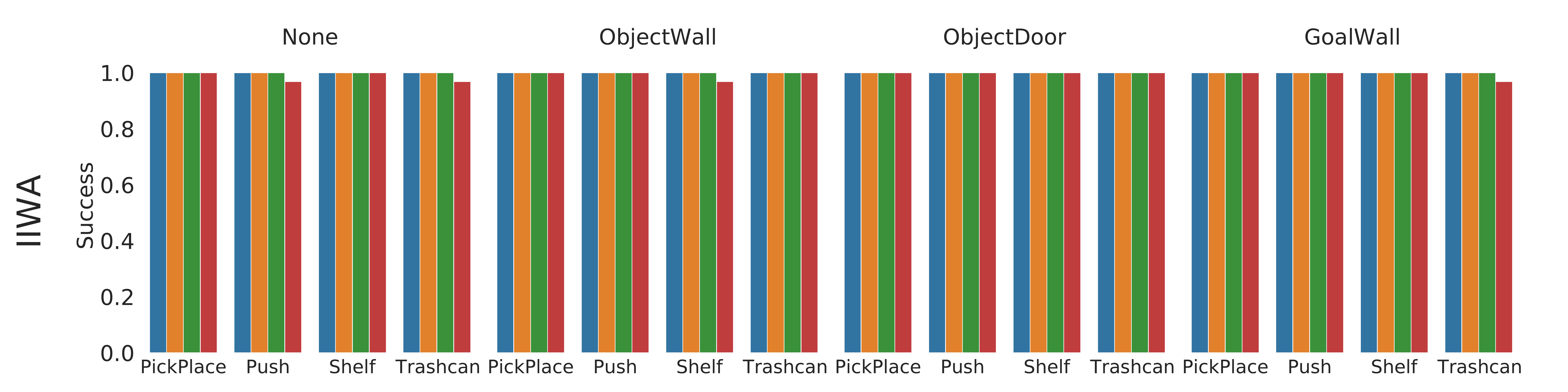}
    \end{subfigure}\\
    \begin{subfigure}[b]{1.\textwidth}
        \centering
            \includegraphics[width=0.75\linewidth, trim={0.3cm 0.0cm 1.5cm 0.8cm},clip]{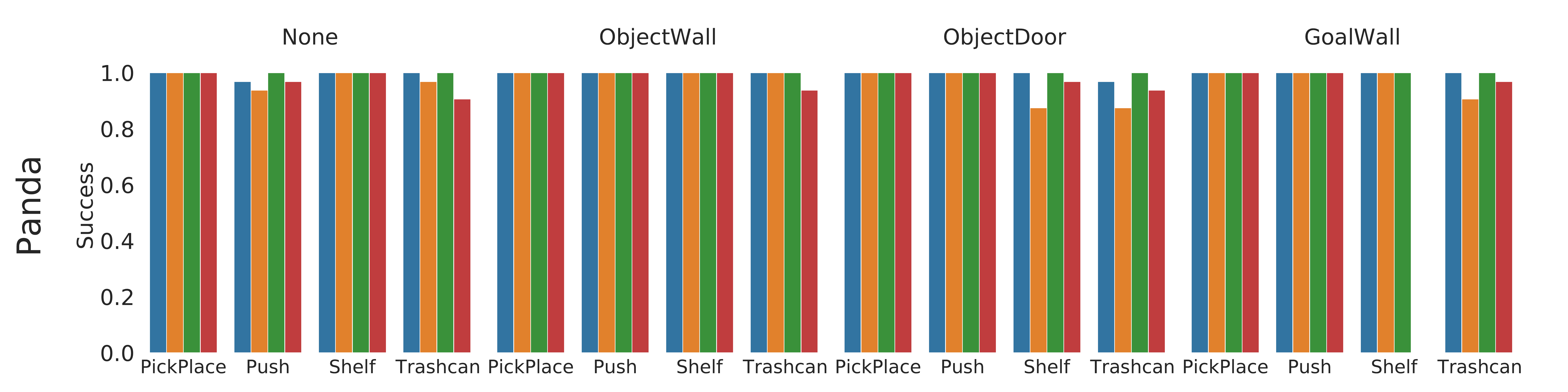}
    \end{subfigure}\\
    \begin{subfigure}[b]{1.\textwidth}
        \centering
            \includegraphics[width=0.75\linewidth, trim={0.3cm 0.0cm 1.5cm 0.8cm},clip]{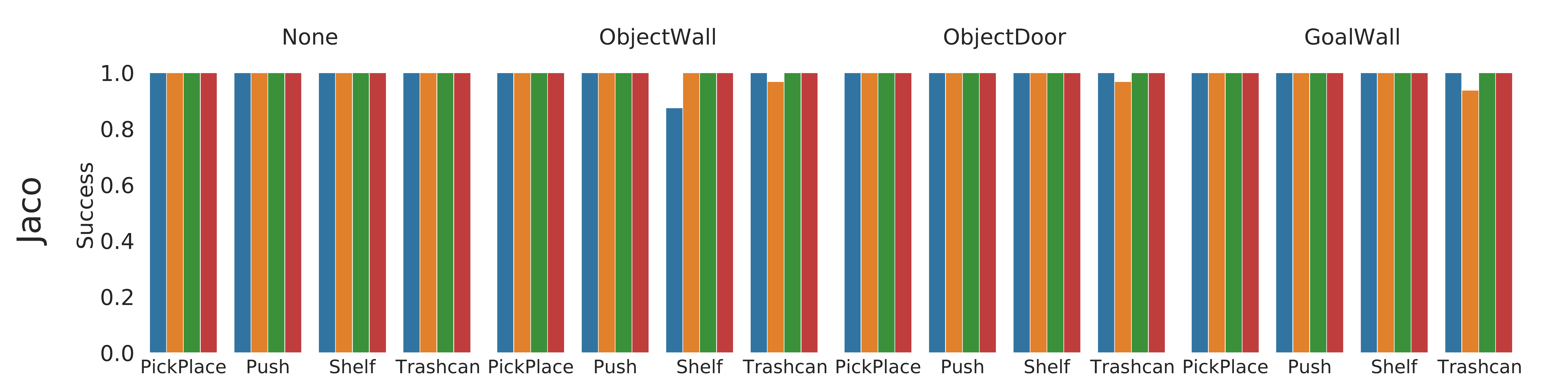}
    \end{subfigure}\\
    \begin{subfigure}[b]{1.\textwidth}
        \centering
            \includegraphics[width=0.75\linewidth, trim={0.3cm 0.0cm 1.5cm 0.8cm},clip]{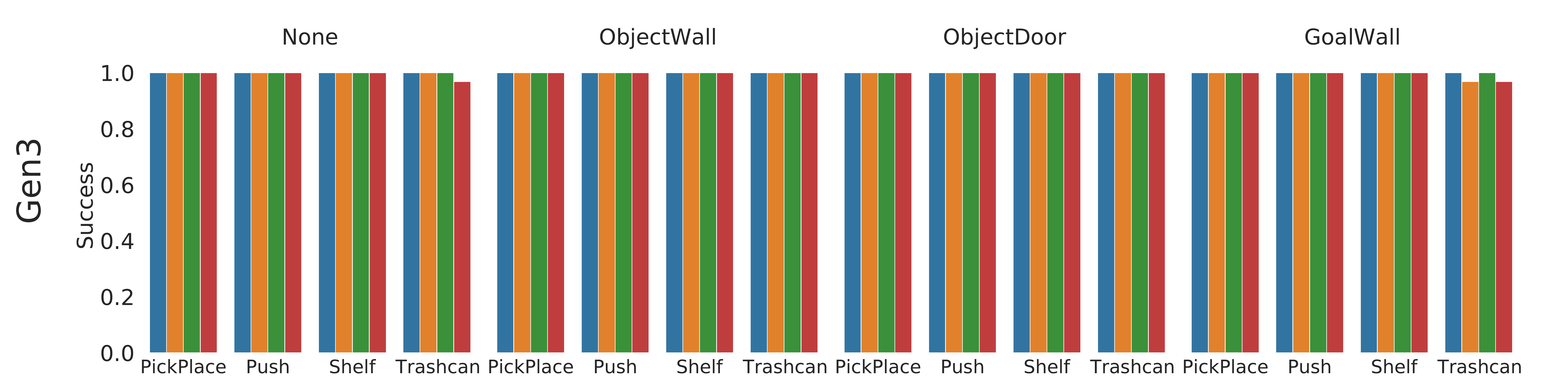}
    \end{subfigure}\\
    \vspace{-1em}
    \caption{Maximum success rate for each task across all trained agents. All but one tasks are solved at least once.}
    \label{fig:maximum_success}
\end{figure}

The combination of elements into the combinatorially many tasks in \benchmark{} raises the question of whether some configurations might lead to tasks that are unsolvable for current RL algorithms, for example by restricting the physical space such that the robot arm cannot fulfill a task objective. In order to validate that the vast majority of tasks in \benchmark{} are solvable, we compute the maximum success over all trained models for every task and visualize it in Figure~\ref{fig:maximum_success}. For all but one task, at least one agent was able to achieve non-zero success. This corroborates that \benchmark{} provides a feasible set of tasks to study compositionality of current RL methods. 

\end{document}